\documentclass[journal]{IEEEtran}
\usepackage{array}
\usepackage[caption=false,font=normalsize,labelfont=sf,textfont=sf]{subfig}
\usepackage{textcomp}
\usepackage{stfloats}
\usepackage{url}
\usepackage{verbatim}
\usepackage{graphicx}
\usepackage{cite}
\usepackage{tabularx}  
\usepackage{siunitx} % 用于数字对齐
\usepackage[normalem]{ulem}
\usepackage{threeparttable}
\usepackage{multirow}%
\usepackage{amsmath,amssymb,amsfonts}%
\usepackage{amsthm}%
\usepackage{mathrsfs}%
\usepackage{xcolor}%
\usepackage{manyfoot}%
\usepackage{booktabs}%
\usepackage{algpseudocode}%
\usepackage{listings}%

\begin{document}

\title{DuSS-PET: Self-Supervised Low-Dose PET Reconstruction Using Pseudo-Dual-Domain Consistency Without Paired Data}

\author{
Shuijin~Lin, 
Yuzhu~He,
Yongbo~Wang,
Jianhua~Ma,
and~Fan~Wang
\thanks{This work did not involve human subjects or animals in its research.}%
\thanks{Shuijin~Lin, Yongbo~Wang, Jianhua~Ma, and Fan~Wang are with the Key Laboratory of Biomedical Information Engineering of Ministry of Education, School of Life Science and Technology, Xi’an Jiaotong University, Xi’an 710049, China, and also with the Research Center for Intelligent Medical Equipment and Devices, Xi’an Jiaotong University, Xi’an 710049, China (e-mail: shuijinlin@stu.xjtu.edu.cn; wyb081@xjtu.edu.cn; jhma@xjtu.edu.cn; fan.wang@xjtu.edu.cn).}%
\thanks{Yuzhu~He is with the School of Mathematics and Statistics, Xi’an Jiaotong University, Xi’an 710049, China, and also with the Research Center for Intelligent Medical Equipment and Devices, Xi’an Jiaotong University, Xi’an 710049, China (e-mail: yuzhu.he@stu.xjtu.edu.cn).}%
\thanks{This work was supported by the National Natural Science Foundation of China (Grant No.~T2522028 and Grant No.~12326616), Fundamental and Interdisciplinary Disciplines Breakthrough Plan of the Ministry of Education of China (No.~JYB2025XDXM101), the Natural Science Basic Research Program of Shaanxi Province, China (Grant No.~2024JC-TBZC-09), the Shaanxi Provincial Key Industrial Innovation Chain Project (Grant No.~2024SF-ZDCYL-02-10), and the Tianyuan Fund for Mathematics of the National Natural Science Foundation of China (Grant No.~12426105). \textit{(Corresponding authors: Fan Wang; Jianhua Ma.)}}%
}

% arXiv author accepted manuscript: publication metadata only.
% IEEE Author Center, Post-Publication Policies, Accepted article.
\IEEEaftertitletext{%
  \begin{center}
  \begin{minipage}{\textwidth}
  \centering
  {\bfseries\normalsize Author Accepted Manuscript\par}
  \smallskip
  {\small IEEE Transactions on Radiation and Plasma Medical Sciences, 2026\par}
  {\small DOI: \url{https://doi.org/10.1109/TRPMS.2026.3696317}\par}
  \smallskip
  {\footnotesize\raggedright
  \textcopyright\ 2026 IEEE. Personal use of this material is permitted.
  Permission from IEEE must be obtained for all other uses, in any current
  or future media, including reprinting/republishing this material for
  advertising or promotional purposes, creating new collective works, for
  resale or redistribution to servers or lists, or reuse of any copyrighted
  component of this work in other works.\par}
  \end{minipage}
  \end{center}
  \vspace{0.5\baselineskip}
}

\maketitle

\begin{abstract}
Positron emission tomography (PET) is pivotal in medicine and healthcare, but requires high radiation doses for quality imaging. Although deep learning has shown promise for low-dose PET reconstruction/enhancement, existing methods primarily rely on fully supervised learning, which requires paired low- and full-dose scans that are clinically impractical to obtain. To overcome this fundamental limitation, we propose \textbf{DuSS-PET}, a novel self-supervised PET reconstruction framework that operates without any paired low- and full-dose data. Our method is built upon a computational Noisier2Noise (Nr2N) paradigm, featuring a Bernoulli corruption strategy that serves as an image-space approximation to emulate low-count noise patterns in image-derived pseudo-sinogram representations. The DuSS-PET framework integrates three core components: an adaptive Swin transformer for pseudo-sinogram restoration, a plug-and-play module for image enhancement, and a consistency-constrained SSL strategy that enforces agreement between the pseudo-sinogram and image representations. Extensive experiments on multi-center datasets demonstrate that DuSS-PET achieves superior reconstruction performance, outperforming state-of-the-art SSL methods and achieving comparable or even better results than fully supervised approaches across various dose levels. In particular, it exhibits exceptional generalization across different scanner vendors. This work is among the first self-supervised low-dose PET reconstruction framework that enforces consistency between reconstructed images and pseudo-sinograms, where the pseudo-sinogram generated in the image space serves as a solid computational consistency constraint for enhanced imaging without full-dose training scans. The source code will be publicly available at \url{https://github.com/ladderlab-xjtu/DuSS-PET}.
\end{abstract}


\section{Introduction}

\IEEEPARstart{P}{ositron} emission tomography (PET) is a cornerstone of nuclear medicine, providing unparalleled sensitivity for in \textit{vivo} functional imaging and metabolic assessment~\cite{apostolovaImagingTumorMetabolism2016}. This capability is critical for detecting abnormalities often indiscernible in anatomical modalities like CT or MRI, solidifying PET's role in healthcare from diagnosis to treatment response evaluation~\cite{zhangComparisonWholebody18FFDG2023,spick18FFDGPETCT2016}. However, achieving high-fidelity PET imaging traditionally relies on elevated radiation doses to ensure sufficient photon-count statistics, which raises concerns regarding patient safety~\cite{gonzalez-montoroAdvancesDetectorInstrumentation2022}.

Low-dose PET protocols aim to mitigate these risks by reducing radiotracer activity. This reduction, however, directly diminishes photon-count statistics, inevitably degrading image quality through heightened noise and a poor signal-to-noise ratio (SNR)~\cite{yanMethodAssessImage2016,watsonOptimizingInjectedDose2005}. Consequently, anomaly detectability and quantitative accuracy are severely compromised, undermining the clinical utility of the scan~\cite{zhaoImageQualityLesion2021}. This critical trade-off establishes the urgent need for advanced reconstruction techniques that can generate high-fidelity images from low-dose acquisitions.

Beyond traditional techniques like filtered back-projection (FBP)~\cite{ramachandran1971three} and ordered-subset expectation maximization (OSEM)~\cite{hudsonAcceleratedImageReconstruction1994}, deep learning has emerged as a powerful paradigm for low-dose PET reconstruction and enhancement, leveraging the exceptional capacity of neural networks for noise suppression and signal recovery~\cite{readerAIPETImage2023}. Existing learning-based approaches can be categorized by their data domain: image-domain enhancement, sinogram-to-image reconstruction, and holistic dual-domain reconstruction. Image-domain methods operate on initially reconstructed PET images~\cite{xiangDeepAutocontextConvolutional2017,wangLocalityAdaptiveMultimodality2018,zhouSupervisedLearningCyclegan2020,zeng3DCVTGAN3D2022,shenPETImageDenoising2023}, offering computational efficiency but often propagating inherent reconstruction artifacts without explicitly enforcing the acquisition physics or projection-space consistency of PET data~\cite{readerAIPETImage2023,sanaatProjectionSpaceImplementation2020}. Conversely, sinogram-to-image networks reconstruct directly from raw projection data~\cite{whiteleyDirectPETFullsizeNeural2020,haggstromDeepPET2019,aiREDResidualEstimation2025,liuDeeplearningbasedFrameworkPET2022,dossUtilizingDeepLearning2023,huTransEMResidualSwintransformer2022,mehranianModelbasedDeepLearning2020}, improving quantitative fidelity by modeling photon statistics but sometimes generating anatomically implausible structures due to a lack of explicit image-domain priors. Recent dual-domain frameworks aim to synergize these strengths, jointly refining information between the sinogram and image spaces to correct errors at their source and refine the final results~\cite{cuiTriDoFormerTripleDomainTransformer2023,cuiPriorKnowledgeguidedTripleDomain2024}. Despite their promising performance, a critical limitation unites these fully supervised methods: their dependence on precisely aligned low-dose and full-dose pairs for training. Such paired data are clinically prohibitive to acquire, arising from ethical and safety concerns associated with subjecting patients to repeated radiotracer injections.

To avoid dependency on paired data, self-supervised learning (SSL) strategies have emerged as a promising alternative~\cite{zhaoSelfsupervisedDeepLearning2023,arabiNoisetonoiseModelTraining2022}. Techniques such as Noise-to-Noise (N2N) and its variants learn to denoise by exploiting the statistical structure within the input data itself, eliminating the need for clean target images~\cite{lehtinenNoise2NoiseLearningImage2018a, moranNoisier2noiseLearningDenoise2020,arabiNoisetonoiseModelTraining2022}. However, a significant limitation of these existing SSL approaches is their predominant operation within the image domain. This post-processing paradigm fails to incorporate the critical physical priors inherent to the sinogram domain, such as photon count statistics and the acquisition model. Consequently, their ability to achieve optimal quantitative reconstruction fidelity from low-count measurements remains fundamentally constrained.

\begin{figure*}[t]
	\centering
	\includegraphics[width=\linewidth]{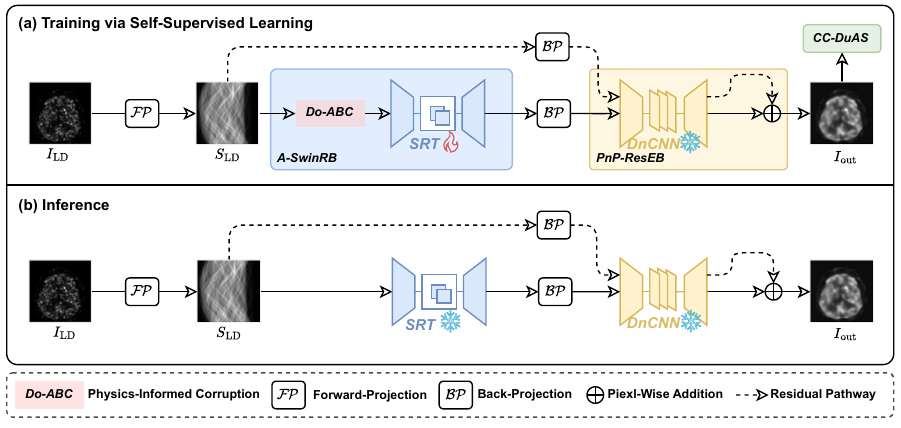}
	\caption{The schematic diagram of DuSS-PET. (a) Training via dual-domain self-supervised learning. A-SwinRB corrects sinograms, where Do-ABC stands for a dose-aware Bernoulli corruption strategy simulating the photon starvation effects of low-dose acquisitions; PnP-ResEB refines image details; and CC-DuAS enhances dual-domain consistency. (b) Inference. }
	\label{fig:net}
\end{figure*}

To overcome the dependency on paired data, we propose a novel self-supervised PET reconstruction framework (\textbf{DuSS-PET}, Fig.~\ref{fig:net}) that extends Noisier-to-Noise (Nr2N) principles from image denoising to the reconstruction setting. Central to our approach is an application-dedicated Bernoulli corruption strategy that approximates low-count noise characteristics associated with low-dose acquisitions in image-derived pseudo-sinograms. Leveraging this tailored corruption process for Nr2N learning, DuSS-PET concurrently refines pseudo-sinograms and reconstructed images through three integrated components: (i) an Adaptive Swin Transformer Restoration Block (A-SwinRB) for pseudo-sinogram correction, (ii) a Plug-and-Play Residual Enhancement Block (PnP-ResEB) for refining spatial details in the image domain, and (iii) a Consistency-Constrained Dual-domain Alignment Strategy (CC-DuAS) to enforce intra- and cross-domain consistency throughout the reconstruction pipeline. Extensive evaluations on multicenter ultra-low-dose PET (UD-PET) benchmarks~\cite{xueCrossscannerCrosstracerDeep2022} demonstrate that DuSS-PET achieves superior PET reconstruction quality in a self-supervised setting. Notably, it delivers performance comparable to, and often surpassing, state-of-the-art fully supervised methods across varying dose levels, while exhibiting significantly stronger cross-vendor generalization, highlighting the effectiveness of enforcing consistency between reconstructed images and pseudo-sinograms.

The principal contributions of this work are three-fold:
\begin{itemize}
\item{We design a Bernoulli sampling-based corruption strategy that injects image-space noise to approximate target low-dose conditions, enabling Nr2N learning for low-dose PET reconstruction without requiring paired full-dose references.}
\item{We improve reconstruction quality and cross-vendor generalization by unifying three self-supervised mechanisms: Nr2N-based restoration of image-derived pseudo-sinograms, Nr2N-based image-domain reconstruction, and a consistency-constrained pseudo-dual-domain regularization that enforces both intra- and inter-domain alignment.}
\item{To our knowledge, DuSS-PET is among the first self-supervised PET reconstruction framework to couple reconstructed images and pseudo-sinograms through an embedded dual-branch consistency mechanism.}
\end{itemize}

The remainder of this paper is organized as follows. First, Section~\ref{sec:2} briefly reviews recent advances in low-dose PET reconstruction, with a particular emphasis on N2N paradigms in medical imaging. Next, Section~\ref{sec:3} introduces the proposed DuSS-PET framework by outlining its motivation and core modules. Subsequently, Section~\ref{sec:4} details the experimental setup, datasets, and ablation studies. Thereafter, Section~\ref{sec:5} discusses the key design choices and their implications. Finally, Section~\ref{sec:6} concludes the paper and points to future directions.

\section{Related work} \label{sec:2}

\subsection{Low-Dose PET Reconstruction and Enhancement}
The evolution of PET reconstruction methodologies has progressed from classical algorithms to modern deep learning paradigms. Classical methods are primarily divided into analytical techniques~\cite{readerAIPETImage2023}, such as FBP, and iterative statistical methods, such as OSEM. While iterative methods improve quantitative accuracy over FBP by modeling the Poisson statistics of photon counting, they remain computationally intensive and often produce noisy images under low-count conditions~\cite{readerAIPETImage2023}.

Driven by these limitations, deep learning approaches have emerged and can be categorized into three groups based on their data processing domain~\cite{fallahpoorDeepLearningTechniques2024,hashimotoDeepLearningBasedPET2024}:
\begin{itemize}
\item{Image-domain post-processing methods, such as LA-GAN~\cite{wangLocalityAdaptiveMultimodality2018} and CycleWGAN~\cite{zhouSupervisedLearningCyclegan2020}, take an initially reconstructed low-dose image as input and enhance its quality. These approaches are computationally efficient but are inherently limited by the artifacts and potential information loss from the initial reconstruction step, often neglecting the underlying physics of PET acquisition.}
\item{Sinogram-to-image reconstruction methods learn a direct mapping from the raw sinogram domain to the image domain. Pioneering works like DeepPET~\cite{haggstromDeepPET2019} and DirectPET~\cite{whiteleyDirectPETFullsizeNeural2020} demonstrated the feasibility of this end-to-end approach. Most recently, the Residual Estimation Diffusion (RED) framework~\cite{aiREDResidualEstimation2025} has established a new state-of-the-art in this category. It employs a diffusion process to iteratively refine residual mappings within the sinogram, significantly improving reconstruction reliability. By leveraging raw photon statistics, these methods can achieve high quantitative fidelity. However, operating primarily in the sinogram domain can sometimes lead to anatomically implausible structures due to a lack of strong image-domain spatial constraints.}
\item{Dual-domain reconstruction methods have recently been proposed to bridge this gap by synergistically processing both sinogram and image data. Frameworks like TriDo-Former~\cite{cuiTriDoFormerTripleDomainTransformer2023} and PK-TriDo~\cite{cuiPriorKnowledgeguidedTripleDomain2024} use triple-domain transformers to enforce cross-domain consistency, where explicit inverse Radon operators are used to transform representations across domains while avoiding unnecessary model complexity. 
These holistic approaches represent a strong recent line of supervised learning for this task.}
\end{itemize}

\begin{figure*}[t]
	\centering
	\includegraphics[width=0.98\textwidth]{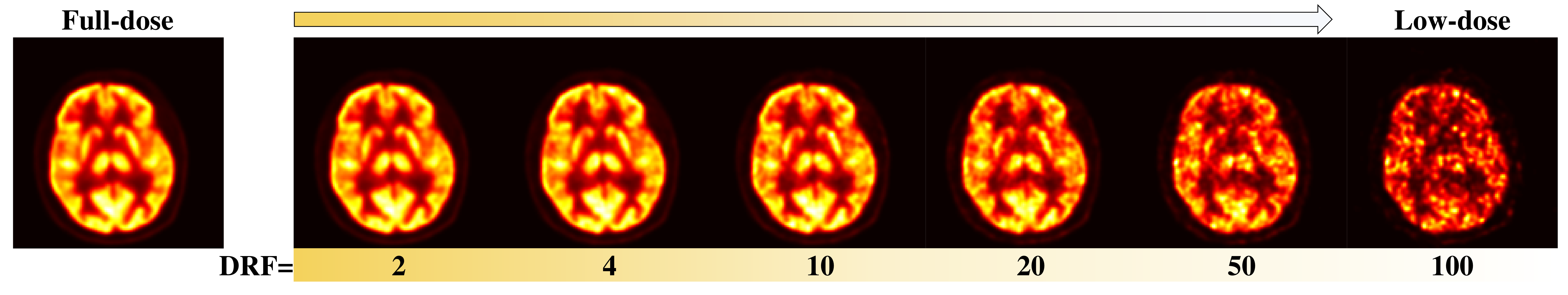}
	\caption{Visualization of full-dose PET image to low-dose images with different dose reduction factors, $\mathrm{DRF}=2, 4, 10, 20, 50, 100$.}
	\label{fig:dose-visual}
\end{figure*}

\begin{figure*}
	\centering
	\includegraphics[width=0.98\textwidth]{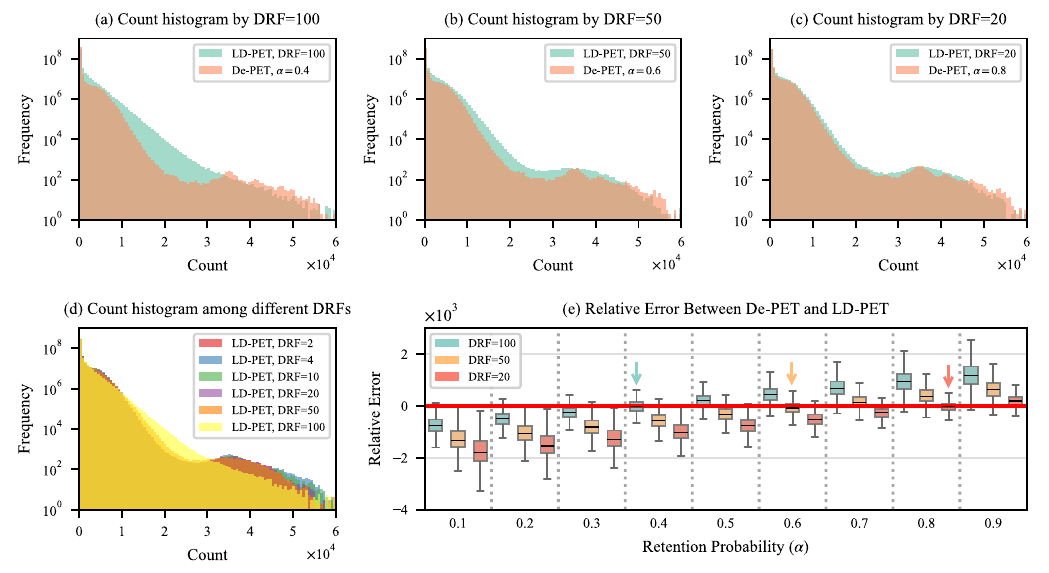}
	\caption{Comparison between LD-PET with different DRFs and De-PET with different Bernoulli retention probabilities $\alpha$. (a)-(c) indicate histograms of voxel counts in the brain region by $\mathrm{DRF}=100, 50, 20$, (d) supplements histograms of LD-PET in all available $\mathrm{DRFs}$, and (e) shows relative error, calculated as $|Count_{\mathrm{LD}} - Count_{\mathrm{De}}| / Count_{\mathrm{LD}}$, where $Count_{\mathrm{LD}}$ and $Count_{\mathrm{De}}$ denote the voxel count sums of the LD-PET and De-PET images, respectively. Three arrows in (e) mark the selected optimal retention probabilities for the corresponding dose levels.}
	\label{fig:count}
\end{figure*}

However, a critical and shared limitation of most existing supervised PET reconstruction and enhancement methods is their dependence on large datasets of precisely aligned low-dose and full-dose scan pairs, which are clinically difficult to acquire due to ethical concerns over repeated radiation exposure.
To mitigate this data constraint, a few self-supervised and weakly-supervised strategies have been explored. Techniques like MRNN~\cite{arabiNoisetonoiseModelTraining2022} employ an N2N paradigm on image data, while frameworks like MAC~\cite{zhaoSelfsupervisedDeepLearning2023} use mask-based cycling. 

Beyond these representative self-supervised/supervised frameworks, recent PET enhancement studies have also begun to explore more flexible data settings~\cite{zhangGeneralizableUnpairedPET2026, gongPETImageDenoisingDDPM2024}. In particular, unpaired image-domain enhancement has recently been investigated using a two-stage diffusion framework~\cite{zhangGeneralizableUnpairedPET2026}, where pseudo paired data are first synthesized and then used to train a second-stage denoiser. This line of work further reflects the growing interest in reducing the reliance on explicitly paired training data in PET image enhancement.
However, these methods often operate solely in a single domain (typically image), neglecting the fundamental physics of sinogram data acquisition. Consequently, a significant gap remains for a holistic, dual-domain self-supervised method that requires no paired data and effectively marries statistical sinogram processing with anatomical image-domain priors. Our work is designed to bridge this gap.
\subsection{N2N-Based Methods for Medical Imaging}

Deep learning has revolutionized image denoising, with SSL paradigms emerging to bypass the need for clean reference data. Among these, Noise-to-Noise (N2N) established a foundational principle, demonstrating that a network can learn to denoise from pairs of different noisy realizations of the same underlying image, provided their noise is statistically independent~\cite{lehtinenNoise2NoiseLearningImage2018a}. This pivotal work unlocked the potential for training denoising models in the absence of clean ground truth.

The core N2N concept has since been extended into more flexible and practical frameworks. Noise-to-Void (N2V) and Noise-to-Self (N2S) moved to a fully blind, single-image setting by learning from the spatial context around masked pixels~\cite{krullNoise2voidlearningDenoisingSingle2019,batsonNoise2selfBlindDenoising2019}. Notably, Nr2N bridged them by synthetically adding further noise to a single noisy input to create the required pair for self-supervision~\cite{moranNoisier2noiseLearningDenoise2020}. 

These principles have been successfully adapted for medical image denoising and reconstruction~\cite{bousseReviewLowDoseEmission2024}, where acquiring pristine ground-truth data is often clinically impossible. In MRI, frameworks like SSDU have been shown to align with the Nr2N principle for reconstructing images from under-sampled data~\cite{yamanSelfsupervisedLearningPhysicsguided2020,millardTheoreticalFrameworkSelfsupervised2023}. In PET, studies have demonstrated that N2N can achieve performance comparable to supervised models in terms of quantitative metrics like PSNR ~\cite{yieSelfsupervisedPETDenoising2020}, while sometimes offering superior preservation of local quantitative accuracy~\cite{yieEffectsLossFunctions2024}. Extensions such as NB2NB~\cite{songSelfsupervisedPETImage2023} and MRNN~\cite{arabiNoisetonoiseModelTraining2022} have introduced innovations like neighborhood subsampling and multi-realization training to enhance stability for PET denoising.
More recent unsupervised PET denoising studies have also explored architectural optimization itself, for example by using neural architecture search to identify more suitable backbones for unsupervised PET image denoising~\cite{liNeuralArchitectureSearchUnsupervised2026,xieUnifiedNoiseAwareNetwork2024}. 

However, a critical limitation persists: these methods are predominantly confined to the image domain. They excel at removing generic noise but often neglect the specific physical acquisition model and statistical properties of raw sensor data (e.g., sinograms in PET). 
This domain restriction may limit the correction of degradation originating in the projection domain and can reduce the fidelity of quantitative and structural recovery, ultimately limiting efficacy and generalizability for full reconstruction tasks~\cite{sanaatProjectionSpaceImplementation2020,readerAIPETImage2023}.

\section{Method} \label{sec:3}
\subsection{Motivation and Overview}
Existing self-supervised denoising methods (e.g., N2N, Nr2N), which are typically formulated without explicit physics guidance and rely on generic Gaussian or mixed noise assumptions, demonstrate limited efficacy for low-dose PET reconstruction. These strategies are primarily designed for image-domain denoising under signal-sufficient conditions and therefore struggle to recover the severely degraded structural information in low-dose PET data, often leading to oversmoothed reconstructions with compromised quantitative accuracy~\cite{songNoise2VoidUnsupervisedDenoising2021}. This limitation stems from a fundamental distinction: image denoising aims to suppress noise in over-sufficient observations~\cite{liuPersonalizedDeepLearning2022}, whereas PET reconstruction is an ill-posed inverse problem that must recover missing information from photon-deficient measurements, traditionally guided by acquisition physics and count statistics~\cite{seyyediDeepLearningbasedTechniques2024a}.

To address these challenges, we propose DuSS-PET, a self-supervised PET reconstruction framework that extends the Nr2N principle by incorporating physics-guided corruption and consistency constraints through image-derived pseudo-sinogram representations. Our approach is driven by two core insights:
\begin{itemize}
\item{\textit{Physics-Guided Corruption}: Effective self-supervision for low-dose PET reconstruction requires noise injection strategies that approximate the statistical characteristics of PET acquisition, rather than relying on generic image-domain perturbations. We propose a Dose-Aware Bernoulli Corruption (Do-ABC) mechanism that operates in the pseudo-sinogram space and calibrates synthetic degradation to approximate photon starvation effects at specific dose levels, providing a physics-guided proxy for low-count PET acquisition within a self-supervised learning framework.}
\item{\textit{Dual-Domain Optimization}: Signal degradation manifests as reduced photon counts in the sinogram. Formulating self-supervised objectives solely in the image domain fails to directly constrain errors originating from photon-deficient measurements. Prior studies suggest that sinogram-space processing can better preserve count-related statistics and projection consistency that are critical for accurate PET reconstruction~\cite{sanaatProjectionSpaceImplementation2020,dossUtilizingDeepLearning2023}. Accordingly, we formulate the self-supervised loss in the pseudo-sinogram domain, where low-count characteristics are more explicitly reflected, while using image-domain reconstruction to produce the final PET images.}
\end{itemize}

Guided by these principles, the DuSS-PET framework integrates three complementary components to enable self-supervised PET reconstruction with sinogram-space consistency constraints (Fig.~\ref{fig:net}): \textbf{(i)} A-SwinRB (Adaptive Swin Transformer Restoration Block): a transformer-based network that integrates Do-ABC to perform primary restoration on corrupted pseudo-sinograms; \textbf{(ii)} PnP-ResEB (Plug-and-Play Residual Enhancement Block): a self-supervised image-domain reconstruction module that refines spatial details and preserves anatomical fidelity in the reconstructed PET images; and \textbf{(iii)} CC-DuAS (Consistency-Constrained Dual-domain Alignment Strategy): a self-supervised consistency constraint that enforces agreement between outputs from differently corrupted sinogram paths and aligns sinogram-space predictions with image-domain reconstructions.

Unlike existing self-supervised PET methods, which are typically confined to a single domain~\cite{arabiNoisetonoiseModelTraining2022,zhaoSelfsupervisedDeepLearning2023}, and supervised dual-domain frameworks~\cite{cuiTriDoFormerTripleDomainTransformer2023,cuiPriorKnowledgeguidedTripleDomain2024}, which rely on paired low-/full-dose data, DuSS-PET integrates three self-supervised components within a pseudo-dual-domain framework. In particular, the sinogram restoration network is incorporated into a dual-branch consistency loop (CC-DuAS), coupling image reconstruction with consistency in the pseudo-sinogram domain. By combining pseudo-sinogram self-supervision with image-domain reconstruction under a unified consistency constraint, DuSS-PET enables low-dose PET reconstruction without requiring paired full-dose references.

%\subsection{Sinogram-domain Restore via A-SwinRB} \label{sec:AA}
\subsection{Physics-Guided Sinogram Restoration with A-SwinRB} \label{sec:AA}

\subsubsection{Empirical Analysis of Sinogram Corruption}

Within the N2N self-supervised framework, the model learns to recover a signal from pairs of corrupted realizations. We define three data states for PET: \textbf{(i)} clean, full-dose (FD-PET); \textbf{(ii)} noisy, low-dose (LD-PET); and \textbf{(iii)} noisier, noisy low-dose (NLD-PET). The objective is to train a network to map from NLD-PET to LD-PET, implicitly learning the denoising function from LD-PET to FD-PET.

To construct a realistic corruption model, we first systematically analyzed PET images acquired under multiple Dose Reduction Factors ($\mathrm{DRF}=2, 4, 10, 20, 50, 100$). Qualitative analysis (Fig.~\ref{fig:dose-visual}) reveals that dose reduction induces two primary effects: (i) a global reduction in image intensity accompanied by increased noise, and (ii) a localized degradation of structural integrity, where contiguous high-uptake regions fragment into scattered, "hollow" patches. This suggests that low-dose PET images can be modeled as structurally degraded versions of their full-dose counterparts.

Motivated by this, we propose a Bernoulli masking strategy to simulate this structural degradation, producing a Degraded PET (De-PET) image. We do not claim that Bernoulli masking constitutes a complete physical model of PET acquisition. Instead, it is introduced as an empirically calibrated corruption proxy for self-supervised learning, motivated by its ability to reproduce two prominent image-level degradation patterns observed in low-dose PET: count-related intensity attenuation and structural fragmentation of high-uptake regions.

To validate this approach, we quantitatively compared the voxel-count distributions of real LD-PET and De-PET images across DRFs. 

The resulting voxel-wise histograms show strong distributional alignment between LD-PET and De-PET at $\mathrm{DRF}=20$ ($\alpha=0.8$), $\mathrm{DRF}=50$ ($\alpha=0.6$), and $\mathrm{DRF}=100$ ($\alpha=0.4$) (Fig.~\ref{fig:count}(a)-(c)), confirming that the Bernoulli model effectively captures the statistical properties of real low-dose data at moderate DRFs. However, a discernible discrepancy remains at the ultra-low-dose condition ($\mathrm{DRF}=100$), where noise dominates the signal and the distribution deviates from typical PET characteristics (Fig.~\ref{fig:count}(d)).

Furthermore, for each DRF, we generated De-PET images using a range of Bernoulli retention probabilities and selected the optimal Bernoulli retention probability by jointly examining (i) the alignment of voxel-count histograms within the brain region and (ii) the relative count-sum error with respect to the corresponding LD-PET images. The selected optimal probabilities are marked by arrows in Fig.~\ref{fig:count}(e).

In conclusion, the Dose-Aware Bernoulli Corruption (Do-ABC) strategy, parameterized by the empirically derived optimal $\alpha$ for each DRF, provides a principled and effective method for simulating low-dose PET degradation. This enables a dose-consistent extension of the Nr2N framework for self-supervised PET reconstruction.

\begin{figure*}
	\centering
	\includegraphics[width=\textwidth]{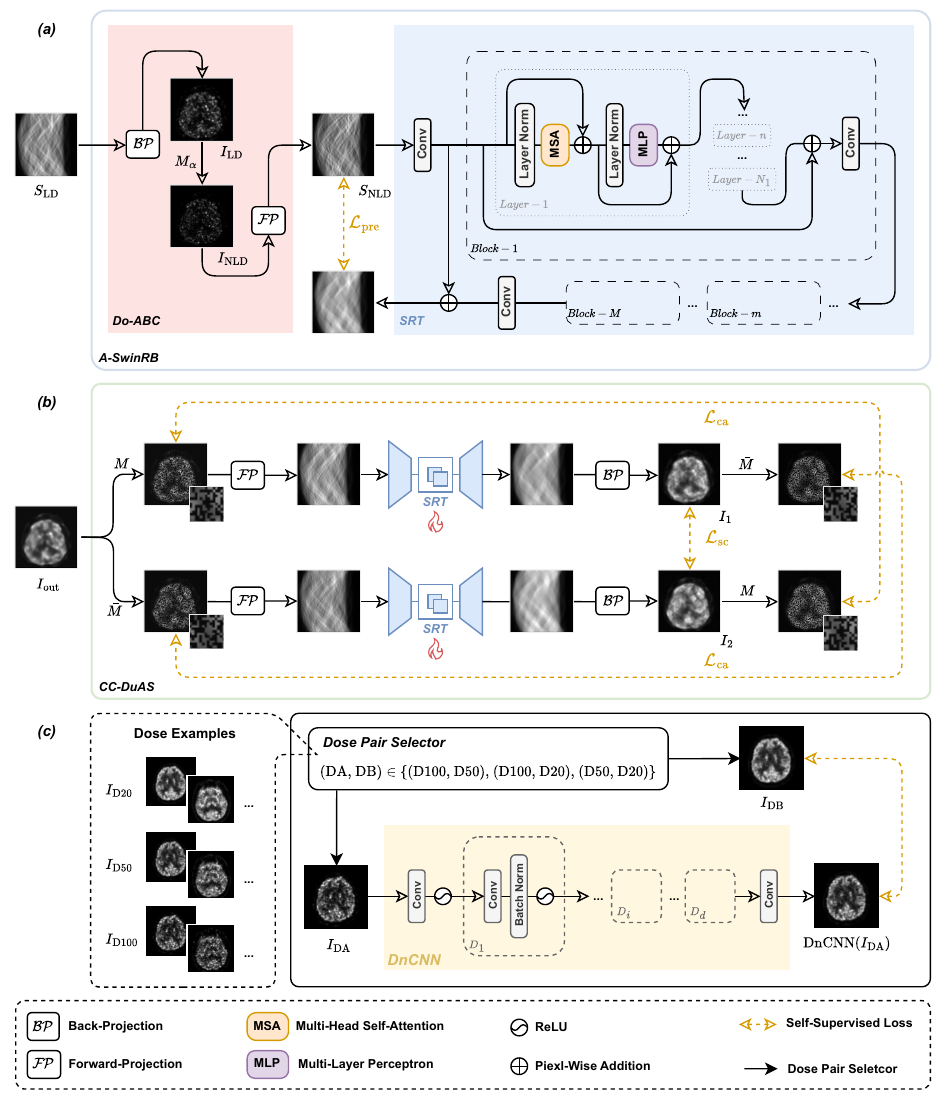}
	\caption{Key components of DuSS-PET. (a) A-SwinRB integrating Do-ABC and SRT. (b) CC-DuAS workflow. (c) Pretraining of DnCNN. $\mathrm{DRF}=20, 50, 100$ are denoted as D20, D50, D100.}
	\label{fig:detail}
\end{figure*}

\subsubsection{Theoretical Formulation of Do-ABC}
Building on the empirical finding that Bernoulli corruption of FD-PET ($I_{\text{FD}}$) yields De-PET images ($I_{\text{De}}$) statistically similar to real LD-PET ($I_{\text{LD}}$), we formalize the degradation process as:
\begin{equation}
    I_{\text{De}} = M_{\alpha_{0}} \odot I_{\text{FD}}, \quad M_{\alpha_{0}} \sim \text{Bernoulli}(\alpha_{0}),
\end{equation}
where $\odot$ denotes element-wise multiplication, and $M_{\alpha_{0}}$ represents a binary masking matrix governed by the retention probability $\alpha_{0}$, where $\alpha_{0} \in [0,1]$ is the retention probability of the Bernoulli mask. It is assumed that $\alpha_{0}$ is optimized for a specific DRF to ensure that $I_{\text{De}}$ aligns with the empirical count-related image statistics of the corresponding $I_{\text{LD}}$ image. This optimization is evaluated through voxel-count histogram alignment and the relative count-sum error, as shown in Fig.~\ref{fig:count}.

To create a paired dataset for Nr2N training, as shown in Fig.~\ref{fig:detail}(a), a secondary Bernoulli corruption is applied to an $I_{\text{LD}}$ image to generate a noisier counterpart $I_{\text{NLD}}$:
\begin{equation}
    I_{\text{NLD}} = M_{\alpha} \odot I_{\text{LD}}, \quad M_{\alpha} \sim \text{Bernoulli}(\alpha),
\end{equation}
where $\alpha \in [0,1]$ is the retention probability of the second corruption.

Given the established approximation $I_{\text{De}} \approx I_{\text{LD}}$, we directly treat $I_{\text{LD}}$ as the noisy target. Thus, $I_{\text{FD}}$, $I_{\text{LD}}$, and $I_{\text{NLD}}$ serve as the clean, noisy, and noisier images, respectively, within the Nr2N framework. Its theoretical foundation provides the following conditional expectation estimator:

Since the original Nr2N derivation for multiplicative Bernoulli noise is expressed in terms of masking-to-zero probabilities $p$ and $q$ rather than retention probabilities, the correspondence in our notation is $p=1-\alpha_0$ and $q=1-\alpha.$

Therefore, following the Nr2N formulation, the conditional expectation estimator can be written as:

\begin{equation}
    \mathbb{E}[I_{\text{FD}} \mid I_{\text{NLD}}] = \frac{\mathbb{E}[I_{\text{LD}} \mid I_{\text{NLD}}] - k \cdot I_{\text{NLD}}}{1 - k},
\label{Eq:fd.nld}
\end{equation}
where $k = (1-\alpha_{0}) / (1 - \alpha \cdot \alpha_{0})$.

Crucially, we migrate this formulation to the sinogram domain to exploit the projection-space representation, where count-related degradation patterns are more explicitly structured. Throughout this work, the term ``sinogram'' refers exclusively to \emph{image-derived pseudo-sinograms} generated by applying a 2D parallel-beam Radon transform to reconstructed PET images, rather than to raw measured projections acquired by the PET scanner. The implications and limitations of this simplified forward model are discussed in Section~\ref{sec:limit}.

Since the forward-projection $\mathcal{FP}(\cdot)$ is a linear operator~\cite{hansenDiscreteInverseProblems2010}, the relationship holds under the transformation:
\begin{equation}
    \mathbb{E}[S_{\text{FD}} \mid S_{\text{NLD}}] = \frac{\mathbb{E}[S_{\text{LD}} \mid S_{\text{NLD}}] - k \cdot S_{\text{NLD}}}{1 - k},
\end{equation}
where $k$ is consistent with that in Eq.~\eqref{Eq:fd.nld}, $S_{\text{FD}} = \mathcal{FP}(I_{\text{FD}})$, $S_{\text{LD}} = \mathcal{FP}(I_{\text{LD}})$, and $S_{\text{NLD}} = \mathcal{FP}(I_{\text{NLD}})$. A detailed derivation is provided in Appendix A.
This estimator allows a network trained on pairs of ($S_{\text{LD}}$, $S_{\text{NLD}}$) to asymptotically learn the mapping from a noisy input to the clean target $S_{\text{FD}}$ despite never being trained with clean data.

\subsubsection{Architecture Design of A-SwinRB}

The mapping between image and sinogram domains in PET reconstruction is inherently asymmetric. While a single image pixel corresponds to a sinusoidal trace in the sinogram, its reconstruction requires integrating information from a broad region of the sinogram~\cite{haggstromDeepPET2019}. This relationship therefore demands a network capable of capturing both local contextual information and long-range dependencies.
To meet these requirements, we design the A-SwinRB module based on the Swin Transformer architecture~\cite{liangSwinirImageRestoration2021}, referred to as the Swin Restoration Transformer (SRT). Its core mechanism, namely shifted window attention, enables efficient multi-scale spatial feature aggregation and position-aware representation learning within the sinogram space. This design allows the SRT to effectively model the global contextual dependencies required for accurate sinogram restoration, while preserving the sensitivity to recover fine-grained structural patterns embedded in severely degraded projections.

\subsection{Plug-and-Play Image Enhancement with PnP-ResEB}
The A-SwinRB module (Sec.~\ref{sec:AA}) primarily operates in the pseudo-sinogram domain to address photon starvation at their origin. To further refine the reconstructed output in the image domain and incorporate essential anatomical priors, we introduce the Plug-and-Play Residual Enhancement Block (PnP-ResEB). This module is a pre-trained image-domain denoiser that refines structural integrity and detail fidelity of the reconstructed PET images without requiring any full-dose supervision.

As shown in Fig.~\ref{fig:detail}(c), PnP-ResEB is pre-trained using an empirically validated Nr2N strategy on real low-dose PET images. Specifically, inspired by~\cite{yieEffectsLossFunctions2024}, it leverages low-dose images acquired at different DRFs to form `noisier-noisy' pairs (e.g., $\mathrm{DRF}=100\rightarrow\mathrm{DRF}=50$), eliminating the need for synthetic corruption and ensuring that the learned priors originate from authentic low-count noise distributions.

Architecturally, PnP-ResEB employs a lightweight backbone, DnCNN~\cite{zhangGaussianDenoiserResidual2017}, to facilitate efficient integration into the reconstruction pipeline. A key design feature is a dedicated \textit{residual pathway} that bypasses the secondary corruption process. This pathway forwards the original, uncorrupted low-dose input directly to the network, preserving critical structural information that would otherwise be degraded and enabling more effective residual learning.

Overall, PnP-ResEB provides three distinct benefits:
\begin{itemize}
\item \textit{Strong Generalization}: Its self-supervised formulation and training on real noise distributions support robust cross-domain and cross-scanner generalization.
\item \textit{Anatomical Prior Injection}: It provides strong image-domain regularization, refining outputs from A-SwinRB to improve structural fidelity and anatomical consistency in the reconstructed images
\item \textit{Multi-Dose Adaptability}: Training on pairs with varying DRFs (e.g., $100\rightarrow50$, $50\to20$) equips the module to accommodate a wide range of degradation levels.
\end{itemize}

Critically, PnP-ResEB is pre-trained on a dataset entirely disjoint from the evaluation data. Each training pair comprises two low-dose realizations of the \emph{same} subject, both retrospectively undersampled from the same list-mode acquisition at different dose reduction factors---not low-/full-dose data from separate scans. This construction is structurally equivalent to the Noisier2Noise paradigm~\cite{moranNoisier2noiseLearningDenoise2020}, under which no full-dose ground-truth is accessed at any stage, thereby preserving the integrity of the overall self-supervised reconstruction paradigm.

\subsection{Dual-Domain Consistency Alignment with CC-DuAS}
\label{sec:CC}

To jointly optimize projection consistency and anatomical plausibility in a self-supervised manner, we introduce a Consistency-Constrained Dual-Domain Alignment Strategy (CC-DuAS), inspired by the consistency-driven design principles in~\cite{zhouDSFormerDualdomainSelfsupervised2023}. This strategy (illustrated in Fig.~\ref{fig:detail}(b)) operates through two complementary consistency mechanisms:

(i) \textit{Self-Consistency Enforcement}: Two shared-weight SRT branches process complementary, Bernoulli-masked versions of the reconstructed image ${I_{\text{out}}}$ (i.e., the output of the A-SwinRB and PnP-ResEB pipeline), generating two intermediate reconstructions $I_1$ and $I_2$:
\begin{equation}
\begin{aligned}
    I_1 &= \mathcal{BP}\left( \text{SRT}\left( \mathcal{FP}\left( M \odot I_{\text{out}} \right) \right) \right), \\
    I_2 &= \mathcal{BP}\left( \text{SRT}\left( \mathcal{FP}\left( \bar{M} \odot I_{\text{out}} \right) \right) \right),
\end{aligned}
\label{Eq:CSM}
\end{equation}
where $\mathcal{FP}(\cdot)$ and $\mathcal{BP}(\cdot)$ denote the Radon and inverse Radon (FBP) transforms, respectively, and $\bar{M}=1-M$. A self-consistency loss ($\mathcal{L}_{\text{sc}}$) enforces agreement between these two predictions, encouraging the model to produce stable reconstructions under different stochastic patterns:
\begin{equation}
\mathcal{L}_{\text{sc}} = \underbrace{||I_1 - I_2||_1}_{\text{self-consistency loss}}.
\label{Eq:lsc}
\end{equation}

(ii) \textit{Cross-Domain Alignment}: A composite cross-alignment regularization ($\mathcal{L}_{\text{ca}}$) links the image domain with its pseudo-sinogram domain. It enforces consistency between the final output $I_{\text{out}}$ and the intermediate masked reconstructions by the SRT branches, thereby regularizing both projection-space consistency and image-domain fidelity:
\begin{equation}
\mathcal{L}_{\text{ca}} =  \underbrace{||\bar{M} \odot I_{\text{out}} - \bar{M} \odot I_1||_1}_{\text{masked loss by $\bar{M}$}} 
+ \underbrace{||M \odot I_{\text{out}} - M \odot I_2||_1}_{\text{masked loss by $M$}}.
\label{Eq:lca}
\end{equation}

Collectively, $\mathcal{L}_{\text{sc}}$ and $\mathcal{L}_{\text{ca}}$ form the CC-DuAS objective, which enforces intra-representation consistency and cross-representation alignment without requiring ground-truth data.

Different from prior dual-domain or physics-consistency methods, which often impose separate constraints in the image and projection domains~\cite{aiREDResidualEstimation2025,cuiPriorKnowledgeguidedTripleDomain2024}, the proposed CC-DuAS incorporates sinogram transformation directly into the reconstruction consistency objective. Specifically, two masked versions of the reconstructed image, generated using complementary Bernoulli masks ($M$ and $\bar{M}$), are propagated through the pseudo-sinogram transformation pipeline ($\mathcal{FP} \rightarrow \text{SRT} \rightarrow \mathcal{BP}$) to produce two correlated reconstructions. CC-DuAS then enforces (1)~\emph{self-consistency} (Eq.~(\ref{Eq:lsc})), which encourages agreement between the two outputs under complementary masking, and (2)~\emph{cross-domain consistency} (Eq.~(\ref{Eq:lca})), which aligns the reconstructed image with its transformed counterpart after pseudo-sinogram domain processing. These constraints together form a consistency loop between the image domain and the pseudo-sinogram domain.
\subsection{Training and Inference Workflow}
\label{sec:loss}

The training of the DuSS-PET framework is conducted in two distinct stages to ensure optimization stability and effectively leverage pre-trained image-domain priors:
\begin{enumerate}
    \item[(i)] \textit{Pre-training}: The DnCNN component within the PnP-ResEB module is independently trained in a self-supervised manner on image-domain data.
    \item[(ii)] \textit{Main Training}: The pre-trained DnCNN weights are frozen. Only the SRT within the A-SwinRB module is optimized. The same SRT parameters are shared across both branches of the CC-DuAS strategy to enforce consistent representation learning.
\end{enumerate}

This design offers significant computational efficiency: A single dose-agnostic DnCNN model, benefiting from the strong generalization of the plug-and-play prior, is used across all dose levels. Each dose setting requires training only a lightweight, dedicated SRT. Critically, the datasets used for pre-training and main training are strictly non-overlapping, and no full-dose images are involved at any stage. The pretraining stage follows the Noisier2Noise paradigm, employing pairs of low-dose realizations from the same subject at different DRFs; the main training stage requires no paired data of any kind. Together, both stages constitute a self-supervised learning paradigm.
Notably, the parameters of the pre-trained DnCNN are frozen during all comparative experiments to ensure a fair and consistent evaluation across different methods.

The training objective $\mathcal{L}_{\text{total}}$ integrates losses from all components:
\begin{equation}
\mathcal{L}_{\text{total}}=\mathcal{L}_\text{pre}+\mathcal{L}_{\text{sc}}+\mathcal{L}_{\text{ca}}.
\end{equation}
The primary sinogram-domain loss $\mathcal{L}_\text{pre}$ enforces projection-space consistency and smoothness:
\begin{equation}
\begin{aligned}
\mathcal{L}_{\text{pre}} = 
&\underbrace{||\text{SRT}(S_{\text{NLD}}) - S_{\text{LD}}||_1}_{\text{sinogram reconstruction}} + \beta \cdot \underbrace{TV(\text{SRT}(S_{\text{NLD}}))}_{\text{TV regularization}},
\end{aligned}
\end{equation}
where $S_{\text{NLD}}$ and $S_{\text{LD}}$ denote the noisier and low-dose pseudo-sinograms, respectively, and $TV(\cdot)$ is a total variation regularization term weighted by $\beta$. The CC-DuAS strategy contributes additional consistency losses $\mathcal{L}_{\text{sc}}$ and $\mathcal{L}_{\text{ca}}$ (Eqs.~(\ref{Eq:lsc}),~(\ref{Eq:lca})) to enhance structural stability and cross-domain alignment.

\textit{Inference}: The final high-fidelity output $I_{\text{out}}$ is generated by fusing the enhanced results from both the restored and original sinogram pathways:
\begin{equation}
I_{\text{out}} = \frac{
    \text{DnCNN}\left( \mathcal{BP}\left( \text{SRT}(S_{\text{LD}}) \right) \right) + 
    \text{DnCNN}\left( \mathcal{BP}(S_{\text{LD}}) \right)
}{2}
\label{Eq:result}
\end{equation}
During inference, the secondary corruption is disabled ($\alpha=0$) following the Nr2N framework, and the network processes the original low-dose pseudo-sinogram $S_{\text{LD}}$.

\begin{table*}[!t]
\caption{Reconstruction performance on \textbf{\texttt{UDPET-Main}}. 
Metrics (PSNR$\uparrow$, SSIM$\uparrow$, NMSE$\downarrow$) are reported at $\mathrm{DRF}=20, 50, 100$. 
Each method is annotated with its supervision strategy ("Self-Sup." for self-supervised, "Sup." for supervised), input domain (sinogram or image), and cross-domain info usage.  
Top-performing methods are in \textbf{bold}, second-best are \uline{underlined}.}
\label{tab:qua}
\centering
\small  
\setlength{\tabcolsep}{3pt} % 控制列间距
\begin{tabular}{l l l l l l l l l} % 文本列固定
\toprule
Method & LD-PET & OSEM & N2V & Nr2N & DeepPET & SpachTrans & RED & DuSS-PET \\
\midrule
Cross-Domain & -- & $\times$ & $\times$ & $\times$ & $\times$ & $\times$ & $\times$ & $\checkmark$ \\
Input Domain      & -- & sinogram & image & image & sinogram & image & sinogram & sinogram \\
Supervision       & -- & -- & Self-Sup. & Self-Sup. & Sup. & Sup. & Sup. & Self-Sup. \\
\midrule
\multicolumn{9}{c}{\textbf{$\mathrm{DRF}=20$}} \\
\midrule
PSNR  & \num{24.09} & \num{24.88} & \num{24.35} & \num{27.18} & \num{24.07} & \uline{\num{28.53}} & \textbf{28.69} & \num{28.00} \\
SSIM  & \num{0.6836} & \num{0.7064} & \num{0.7364} & \num{0.7549} & \num{0.8096} & \num{0.7866} & \uline{\num{0.8975}} & \textbf{0.9118} \\
NMSE  & \num{0.0533} & \num{0.0448} & \num{0.0499} & \num{0.0248} & \num{0.0640} & \uline{\num{0.0184}} & \textbf{0.0179} & \num{0.0219} \\
\midrule
\multicolumn{9}{c}{\textbf{$\mathrm{DRF}=50$}} \\
\midrule
PSNR  & \num{19.65} & \num{20.33} & \num{19.86} & \num{23.59} & \num{23.70} & \uline{\num{26.31}} & \textbf{26.39} & \num{25.92} \\
SSIM  & \num{0.5700} & \num{0.6065} & \num{0.6232} & \num{0.7244} & \num{0.7693} & \num{0.7532} & \uline{\num{0.8389}} & \textbf{0.8594} \\
NMSE  & \num{0.1438} & \num{0.1250} & \num{0.1374} & \num{0.0566} & \num{0.0672} & \uline{\num{0.0307}} & \textbf{0.0311} & \num{0.0342} \\
\midrule
\multicolumn{9}{c}{\textbf{$\mathrm{DRF}=100$}} \\
\midrule
PSNR  & \num{15.83} & \num{16.20} & \num{15.89} & \num{20.80} & \num{22.69} & \textbf{24.05} & \uline{\num{23.57}} & \num{23.08} \\
SSIM  & \num{0.4227} & \num{0.4565} & \num{0.4827} & \num{0.6739} & \textbf{0.7829} & \num{0.7286} & \num{0.7062} & \uline{\num{0.7815}} \\
NMSE  & \num{0.3302} & \num{0.3064} & \num{0.3262} & \num{0.1088} & \num{0.0881} & \textbf{0.0537} & \uline{\num{0.0592}} & \num{0.0649} \\
\bottomrule
\end{tabular}
\end{table*}

\section{Experiments and Results} \label{sec:4}

\subsection{Datasets and Preprocessing}
This study utilized three distinct subsets from the publicly available UD-PET repository~\cite{xueCrossscannerCrosstracerDeep2022}, comprising whole-body $^{18}$F-FDG PET scans acquired on two PET scanner models: a United Imaging uEXPLORER and a Siemens Biograph Vision Quadra. All scans were acquired in list-mode, enabling retrospective simulation of low-dose acquisitions at multiple dose reduction factors (DRFs). For computational efficiency and to focus on a high-uptake cerebral region, 25 central axial slices ($128\times128$ resolution) centered on the brain were extracted from each subject.

The datasets were rigorously partitioned to ensure a fair and leak-proof evaluation under a strict SSL paradigm:

\textbf{\texttt{UDPET-Pretrain}} (United Imaging, 2023, $n=664$): This dataset was used exclusively for the self-supervised pre-training of the DnCNN module within PnP-ResEB via the Nr2N strategy. It contains only low-dose PET images; no full-dose data were accessed at any stage. The resulting DnCNN weights were frozen for all subsequent experiments. 

\textbf{\texttt{UDPET-Main}} (United Imaging, 2022, $n=235$): This dataset was used for the main training, validation, and testing of the A-SwinRB module and for benchmarking against all competing methods. It was split into $188/24/23$ subjects for training, validation, and testing, respectively. It shares no subjects with \textbf{\texttt{UDPET-Pretrain}}.

\textbf{\texttt{UDPET-Generalization}} (Siemens, 2021-2022, $n=28$): This dataset, acquired on a different scanner model (Siemens) at a different institution (Bern), was reserved exclusively for zero-shot generalization evaluation. It was excluded from all training, validation, and hyperparameter tuning phases, providing a rigorous assessment of cross-scanner and cross-site robustness.

Notably, \texttt{UDPET-Pretrain} was acquired one year later than \texttt{UDPET-Main}, deliberately introducing a mild distributional shift that serves as an implicit test of the pre-trained DnCNN prior's generalizability across moderate data distribution changes.

\begin{figure*}
	\centering
	\includegraphics[width=\textwidth]{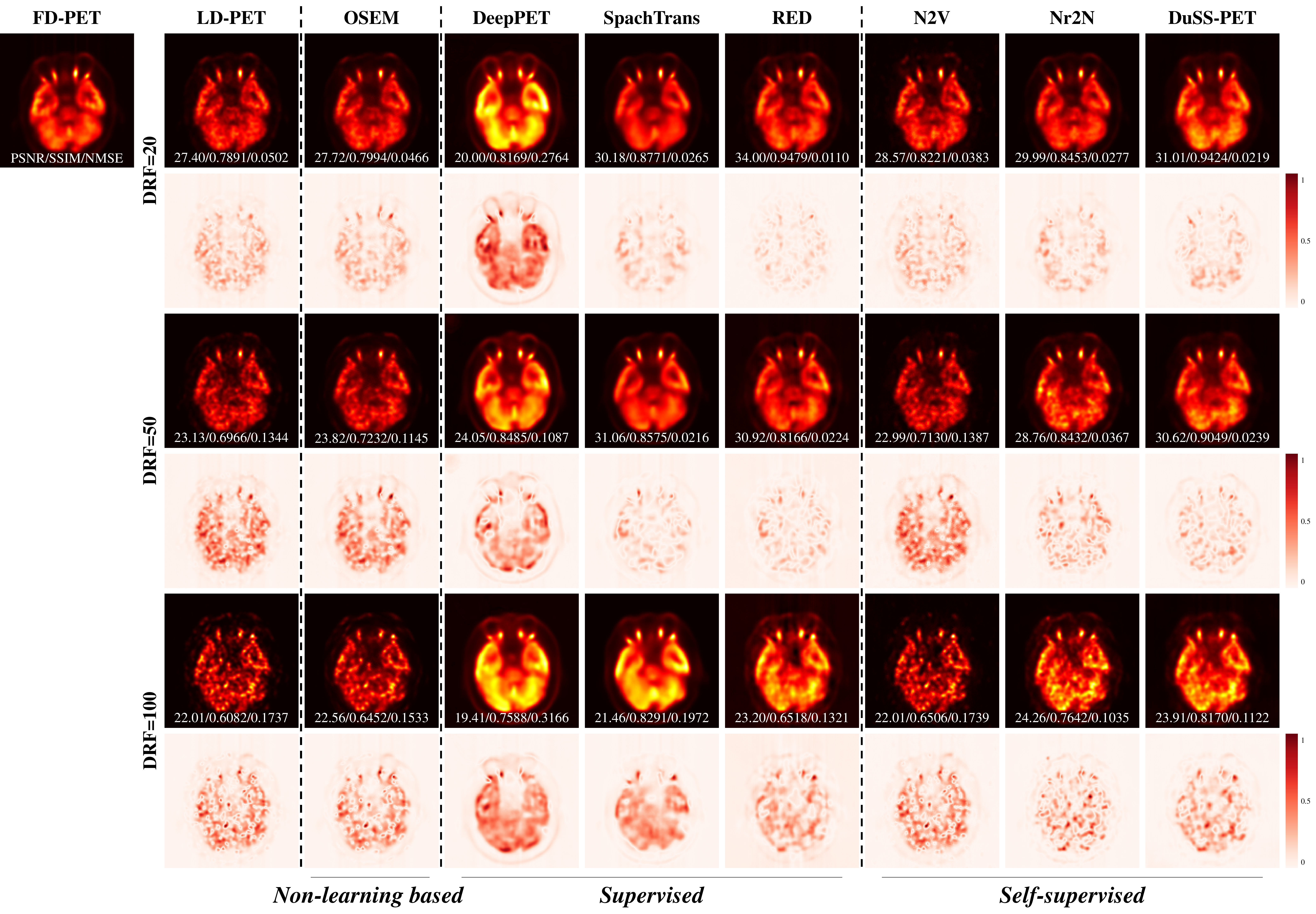}
	\caption{Reconstruction results and corresponding error maps on the \textbf{\texttt{UDPET-Main}} dataset at $\mathrm{DRF}=20, 50, 100$. Methods are grouped as Non-learning based, Supervised, and Self-supervised. PSNR, SSIM, and NMSE are shown below each image. Odd rows: reconstructed PET; even rows: error maps relative to FD-PET.}

	\label{fig:M.E}
\end{figure*}

\subsection{Experimental Setting}
\subsubsection{Implementation Details} 
The DuSS-PET framework was trained in two sequential stages. First, the DnCNN-based PnP-ResEB module was pre-trained using a self-supervised Nr2N paradigm on image-domain PET images. Subsequently, with the DnCNN weights frozen, the main reconstruction framework was optimized for $100$ epochs using Stochastic Gradient Descent (SGD) with a momentum of $0.9$, a batch size of $4$, and an initial learning rate of $1e-4$. The sinogram total variation regularization weight $\beta$ was set to $0.8$. SRT was implemented following the SwinIR architecture~\cite{liangSwinirImageRestoration2021} with a window size of $4\times4$ and depth configuration $[N_1,N_2,N_3]=[2,6,2]$. All models were developed in PyTorch and trained on two NVIDIA RTX 3090 GPUs.

\subsubsection{Compared Methods} 
We evaluated DuSS-PET against six representative methods across three dose reduction factors ($\mathrm{DRF}=20, 50, 100$), encompassing classical, self-supervised, and fully supervised deep learning approaches:
\begin{itemize}
    \item \textbf{OSEM}~\cite{hudsonAcceleratedImageReconstruction1994}: A classical iterative reconstruction baseline, implemented with three iterations and seventeen subsets following widely adopted settings, to reconstruct images from image-derived sinograms under a retrospective reconstruction protocol.
    \item \textbf{N2V}~\cite{songNoise2VoidUnsupervisedDenoising2021}: A self-supervised denoising method. For a fair architectural comparison, we re-implemented it using a Swin Transformer backbone instead of the original U-Net.
    \item \textbf{Nr2N}~\cite{yieEffectsLossFunctions2024}: A self-supervised method based on double-noise corruption, reproduced from its literature description.
    \item \textbf{DeepPET}~\cite{haggstromDeepPET2019}: A fully supervised, end-to-end model mapping sinograms to images.
    \item \textbf{SpachTrans}~\cite{jangSpachTransformerSpatial2024}(TMI, 2024): A supervised transformer-based framework. We adapted its 3D architecture to 2D for our data.
    \item \textbf{RED}~\cite{aiREDResidualEstimation2025} (MedIA, 2025): A recent, state-of-the-art supervised method using a diffusion model for sinogram-domain correction.
\end{itemize}

\subsubsection{Evaluation Protocol}
Following~\cite{cuiPriorKnowledgeguidedTripleDomain2024}, all forward and inverse transformations between the image domain and the image-derived sinogram domain were implemented using the same torch\_radon package~\cite{ronchettiTorchRadonFastDifferentiable2020} across all methods, ensuring consistent and differentiable computational mappings for fair comparison. Critically, N2v, Nr2N, and the proposed DuSS-PET were trained in a strictly self-supervised manner without access to any full-dose ground-truth images. In contrast, all supervised baseline methods (DeepPET, SpachTrans, and RED) were trained using paired low-dose and full-dose data, following their original training protocols.

\subsection{Computational Cost Analysis}
To assess the practical feasibility of deploying DuSS-PET in clinical workflows, we compare the computational cost of all methods in Table~\ref{tab:compute}. All measurements were obtained on a single NVIDIA RTX 3090 GPU (24\,GB).

\begin{table}[htbp]
\small
\centering
\renewcommand{\arraystretch}{1.15}
\begin{threeparttable}
\caption{ Computational cost comparison. Params denotes the total number of learnable parameters. Training time is reported per dose level. Inference time is measured per 2D slice.}
\label{tab:compute}
\begin{tabularx}{\linewidth}{l l r r r}
\toprule
 Method &  Sup. &  Params\,(M) &  Train\,(h) &  Infer.\,(ms) \\
\midrule
 N2V$^{\dag}$   &  Self &  3.79  &  ${\sim}$0.5  &  41.13 \\
 Nr2N           &  Self &  3.36  &  ${\sim}$0.5  &  3.14  \\
 DeepPET        &  Sup. &  63.62 &  ${\sim}$0.1  &  12.19 \\
 SpachTrans     &  Sup. &  3.73  &  ${\sim}$1--2 &  41.19 \\
 RED$^{\ddag}$  &  Sup. &  28.93 &  ${\sim}$24   &  715 \\
 \textbf{DuSS-PET}$^{\S}$ &  \textbf{Self} &  \textbf{4.35} &  $\boldsymbol{{\sim}}$\textbf{40} &  \textbf{${\sim}$47} \\
\bottomrule
\end{tabularx}
\begin{tablenotes}\footnotesize
\item[$\dag$] N2V uses the same SRT backbone as DuSS-PET for fair architectural comparison.
\item[$\ddag$] RED requires sequential training of two sub-networks (REN and DCN) and iterative diffusion sampling (20 reverse steps) at inference, leading to substantially higher inference time.
\item[$\S$] DuSS-PET parameters: SRT (3.79\,M) + DnCNN (0.56\,M, frozen) = 4.35\,M. Inference (${\sim}$47\,ms) includes both reconstruction network and denoiser forward pass, excluding deterministic FP/BP operators shared by all sinogram-based methods. Training time corresponds to the main stage only (100 epochs per DRF, ${\sim}$24\,min/epoch); the DnCNN pre-training is a one-time cost of ${\sim}$0.5\,h shared across all dose levels.
\end{tablenotes}
\end{threeparttable}
\end{table}

As shown in Table~\ref{tab:compute}, DuSS-PET maintains a compact model size of 4.35\,M parameters, which is only 6.8\% of DeepPET (63.62\,M) and 15\% of RED (28.93\,M), comparable to Nr2N (3.36\,M) and SpachTrans (3.73\,M). At inference, DuSS-PET processes each 2D slice in ${\sim}$47\,ms, which is approximately 15$\times$ faster than RED (${\sim}$715\,ms) whose iterative diffusion sampling dominates its computational overhead. While DuSS-PET's main training stage (100 epochs per DRF) requires ${\sim}$40\,h, this reflects the cost of simultaneously optimizing the sinogram restoration network and the dual-domain consistency loop; the image-domain DnCNN prior is pre-trained once (${\sim}$0.5\,h) and reused across all dose levels without retraining. Overall, DuSS-PET achieves a favorable balance between reconstruction quality and computational efficiency, with inference times less and well within the range acceptable for clinical deployment.

\begin{figure*}
	\centering
	\includegraphics[width=\textwidth]{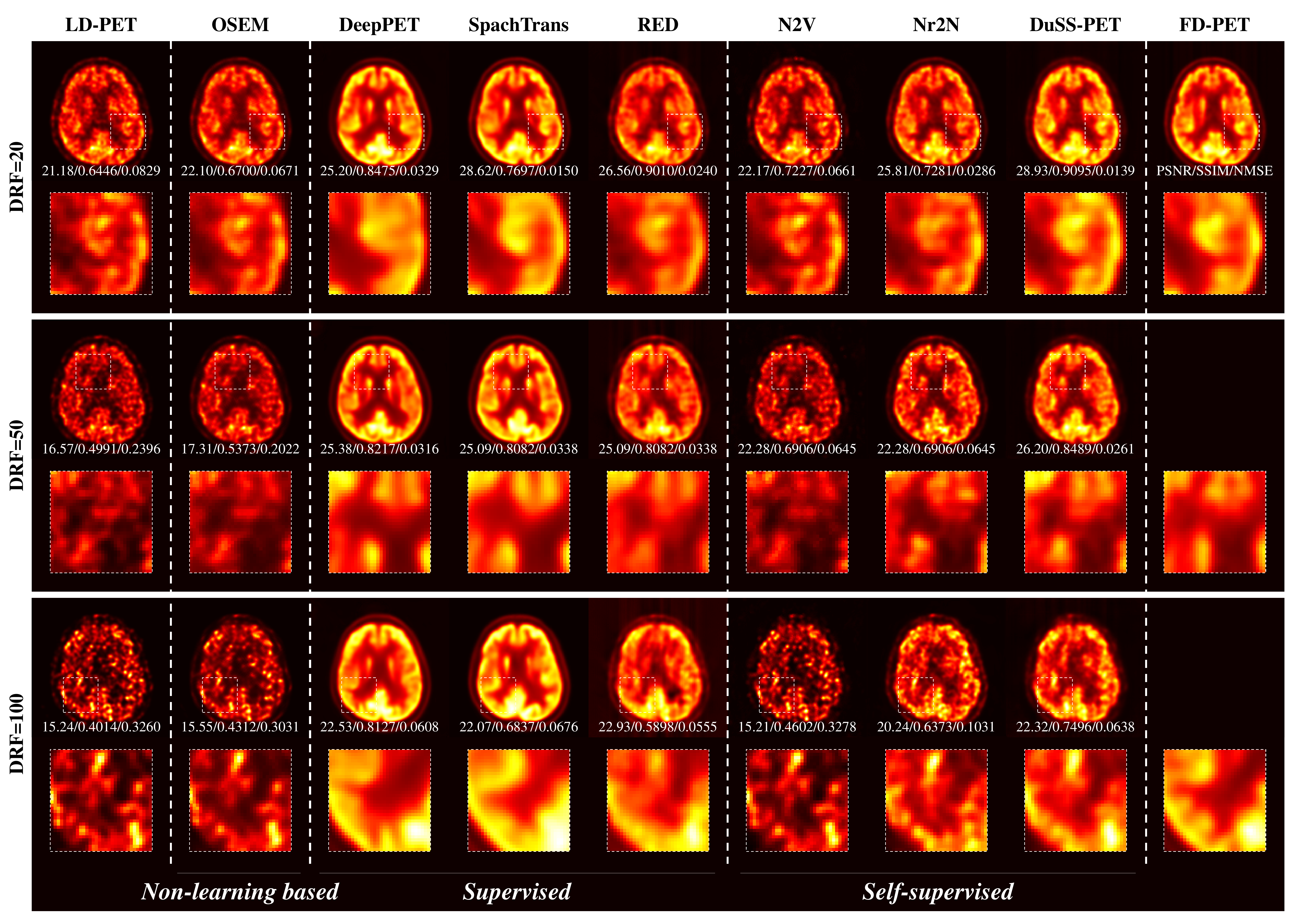}
    \caption{Reconstruction results and corresponding zoom-in details on the \textbf{\texttt{UDPET-Main}} dataset at $\mathrm{DRF}=20, 50, 100$. Methods are grouped as Non-learning based, Supervised, and Self-supervised. PSNR, SSIM, and NMSE are shown below each image. Odd rows: reconstructed PET; even rows: zoom-in views highlighting local structural details.}
    
	\label{fig:M.Z}
\end{figure*}

\subsection{Reconstruction Results Across Varying Dose Levels}

We quantitatively evaluated DuSS-PET on the \textbf{\texttt{UDPET-Main}} test set across three dose reduction factors ($\mathrm{DRF}=20, 50, 100$). As summarized in Table~\ref{tab:qua}, our method consistently achieves superior or competitive performance across all metrics, including Peak Signal-to-Noise Ratio (\textbf{PSNR}), Structural Similarity Index (\textbf{SSIM}), and Normalized Mean Squared Error (\textbf{NMSE}). Key quantitative findings include:
\begin{itemize}
\item DuSS-PET achieves the highest SSIM scores at $\mathrm{DRF}=20, 50$, outperforming other methods by more than $0.02$. At $\mathrm{DRF}=100$, it remains competitive, trailing the top-performing supervised method (DeepPET) by a marginal $0.002$.
\item The performance gap in PSNR between our self-supervised method and the best fully supervised approaches is less than $1$ dB across all dose levels.
\item DuSS-PET maintains low NMSE values, remaining within $0.01$ of the optimal score, which underscores its robustness across varying noise conditions.
\end{itemize}

Qualitative results are presented in Fig.~\ref{fig:M.E} (reconstructions and error maps) and Fig.~\ref{fig:M.Z} (zoomed-in details). Visual inspection reveals that, among self-supervised methods, N2V fails to reliably recover anatomical structures, particularly at high DRFs, while Nr2N exhibits improved but still limited recovery of fine structural details. In contrast, our DuSS-PET successfully reconstructs subtle anatomical features (e.g., hollow regions) that other self-supervised methods miss. Critically, it produces images with visual quality and structural fidelity comparable to state-of-the-art supervised models, despite being trained without any full-dose ground-truth data.

It is worth noting that RED, as a fully supervised diffusion-based method trained with paired full-dose references, achieves higher PSNR and lower NMSE than DuSS-PET on the in-distribution \textbf{\texttt{UDPET-Main}} test set. Such an advantage is expected because access to ground-truth full-dose labels provides a direct pixel-level optimization target that favors fidelity-oriented metrics. DuSS-PET, however, consistently achieves higher SSIM across all dose levels, suggesting better preservation of structural and perceptual information.

These results demonstrate that our dual-domain self-supervised framework not only outperforms other SSL approaches but also narrows the performance gap with fully supervised paradigms on in-distribution data.

\begin{table*}[!t]
\caption{Zero-shot reconstruction performance on \textbf{\texttt{UDPET-Generalization}}. 
Average PSNR, SSIM, and NMSE are reported at $\mathrm{DRF}=20, 50, 100$. 
Annotations indicate supervision strategy, input domain, and cross-domain information, consistent with Table~\ref{tab:qua}. 
Best results are shown in \textbf{bold}, second-best are \uline{underlined}.}
\label{tab:gen}

\centering
\small  
\setlength{\tabcolsep}{3pt} % 控制列间距

\begin{tabular}{l l l l l l l l l}
\toprule
Method & LD-PET & OSEM & N2V & Nr2N & DeepPET & SpachTrans & RED & DuSS-PET \\
\midrule
Cross-Domain & -- & $\times$ & $\times$ & $\times$ & $\times$ & $\times$ & $\times$ & $\checkmark$ \\
Input Domain      & -- & sinogram & image & image & sinogram & image & sinogram & sinogram \\
Supervision       & -- & -- & Self-Sup. & Self-Sup. & Sup. & Sup. & Sup. & Self-Sup. \\
\midrule
\multicolumn{9}{c}{\textbf{$\mathrm{DRF}=20$}} \\
\midrule
PSNR & \num{23.48} & \num{24.97} & \num{24.35} & \num{26.87} & \num{20.26} & \num{25.83} & \textbf{28.49} & \uline{\num{28.46}} \\
SSIM & \num{0.6685} & \num{0.6948} & \num{0.7216} & \num{0.8109} & \num{0.6727} & \num{0.7582} & \uline{\num{0.8882}} & \textbf{0.9183}\\
NMSE & \num{0.0504} & \num{0.0365} & \num{0.0412} & \num{0.0208} & \num{0.1286} & \num{0.0270} & \textbf{0.0145} & \uline{\num{0.0149}} \\
\midrule
\multicolumn{9}{c}{\textbf{$\mathrm{DRF}=50$}} \\
\midrule
PSNR & \num{19.07} & \num{20.29} & \num{19.75} & \num{20.67} & \num{20.28} & \num{24.90} & \uline{\num{26.01}} & \textbf{26.93} \\
SSIM & \num{0.5724} & \num{0.6155} & \num{0.6309} & \num{0.5807} & \num{0.6498} & \num{0.7700} & \uline{\num{0.8270}} & \textbf{0.8718}\\
NMSE & \num{0.1348} & \num{0.1051} & \num{0.1160} & \num{0.0942} & \num{0.1248} & \num{0.0334} & \uline{\num{0.0253}} & \textbf{0.0210} \\
\midrule
\multicolumn{9}{c}{\textbf{$\mathrm{DRF}=100$}} \\
\midrule
PSNR & \num{15.81} & \num{16.71} & \num{16.22} & \num{21.31} & \num{19.67} & \num{23.44} & \uline{\num{23.89}} & \textbf{25.00} \\
SSIM & \num{0.4548} & \num{0.5016} & \num{0.5292} & \num{0.6931} & \num{0.6697} & \uline{\num{0.7547}} & \num{0.7139} & \textbf{0.8310} \\
NMSE & \num{0.2838} & \num{0.2393} & \num{0.2585} & \num{0.0848} & \num{0.1396} & \num{0.0461} & \uline{\num{0.0415}} & \textbf{0.0331} \\
\bottomrule
\end{tabular}
\end{table*}

\begin{figure*}
	\centering
	\includegraphics[width=\textwidth]{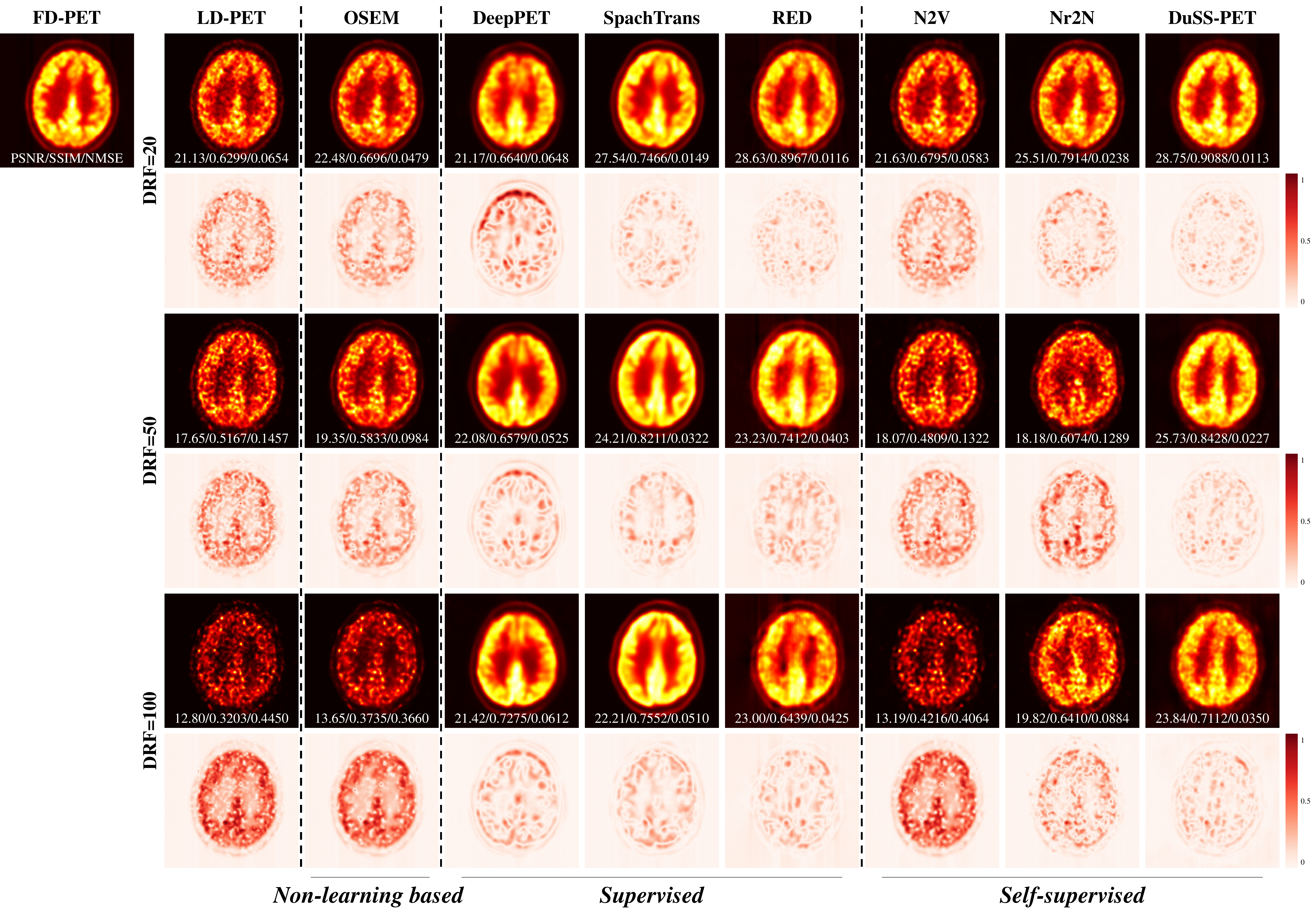}
    \caption{Zero-shot reconstructions and corresponding \textbf{error maps} on the \textbf{\texttt{UDPET-Generalization}} dataset at $\mathrm{DRF}=20, 50, 100$. Methods are grouped as Non-learning based, Supervised, and Self-supervised. PSNR, SSIM, and NMSE are shown below each image. Odd rows: reconstructed PET; even rows: error maps relative to FD-PET.}

	\label{fig:Z.E}
\end{figure*}

\begin{figure*}
	\centering
	\includegraphics[width=\textwidth]{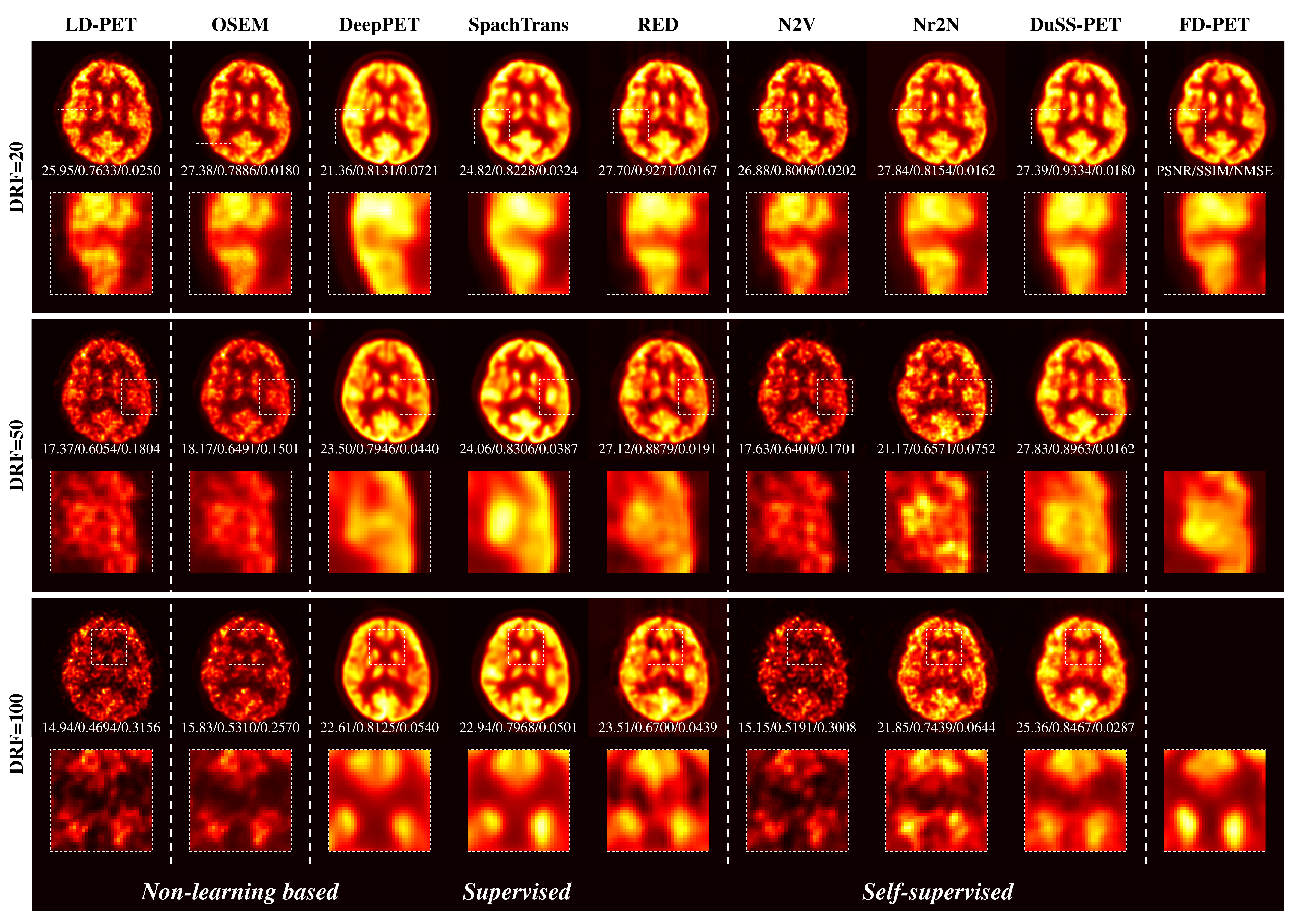}
    \caption{Zero-shot reconstructions and corresponding zoom-in details on the \textbf{\texttt{UDPET-Generalization}} dataset at $\mathrm{DRF}=20, 50, 100$. Methods are grouped as Non-learning based, Supervised, and Self-supervised. PSNR, SSIM, and NMSE are shown below each image. Odd rows: reconstructed PET; even rows: zoom-in views highlighting local structural details.}

	\label{fig:Z.Z}
\end{figure*}

\subsection{Zero-Shot Generalization Across Scanner Vendors}
To evaluate the cross-domain generalization capability of DuSS-PET, we conducted a zero-shot transfer experiment. Models trained on the \textbf{\texttt{UDPET-Main}} dataset (acquired from United Imaging scanners) were directly applied to the held-out \textbf{\texttt{UDPET-Generalization}} dataset (acquired from Siemens scanners) without any fine-tuning.

Quantitative results (Table~\ref{tab:gen}) demonstrate the superior and robust performance of DuSS-PET across all three dose levels ($\mathrm{DRF}=20, 50, 100$). The performance gains of our method become increasingly pronounced as the dose decreases: At $\mathrm{DRF}=20$, DuSS-PET performs on par with the best methods, with a marginal difference of $-0.03$ dB in PSNR and within $3\%$ for SSIM and NMSE; at $\mathrm{DRF}=50$, it achieves significant improvements, i.e., $+0.92$ dB in PSNR, $+6\%$ in SSIM, and $+16\%$ in NMSE; and at $\mathrm{DRF}=100$, the advantage is most substantial: $+1.11 dB$ in PSNR, $+17\%$ in SSIM, and $+24\%$ in NMSE.

This resilience to extreme photon sparsity and domain shift underscores the strong generalization inherent in our self-supervised, dual-domain design. The integrated components (A-SwinRB, PnP-ResEB, and CC-DuAS) collectively enabled robust structural recovery by the proposed self-supervised design with image/sinogram representation consistency.

\textit{In contrast}, fully supervised methods like DeepPET exhibit substantial performance degradation on unseen data, highlighting a key vulnerability of paired-data dependency and the advantage of our approach.

Qualitative results (Fig.~\ref{fig:Z.E} for full reconstructions and error maps; Fig.~\ref{fig:Z.Z} for zoomed details) visually confirm this analysis. DuSS-PET successfully recovers fine anatomical details (e.g., cortical boundaries, deep gray matter structures) that are missed by other self-supervised methods (N2V, Nr2N) and produces markedly lower reconstruction error than all fully supervised baselines.

\begin{table*}[t!]
\small
\centering
\renewcommand{\arraystretch}{1.15}
% \begin{threeparttable}
\caption{Ablation study of core modules in the proposed framework under $\mathrm{DRF}=50$ using PSNR, SSIM, and NMSE.}
\begin{tabularx}{\textwidth}{lXXX}
\toprule
Ablation Setting & PSNR & SSIM & NMSE \\
\midrule
LD-PET & 19.65 & 0.5700 & 0.1438 \\
(i) baseline & 22.02 & 0.8144 & 0.0857 \\
(ii) baseline + PnP-ResEB & 25.59 & 0.8549 & 0.0368 \\
(iii) baseline + CC-DuAS & 23.00 & 0.8256 & 0.0619 \\
(iv) baseline + PnP-ResEB$^{*}$ + CC-DuAS & 25.59 & 0.8529 & 0.0377 \\
(v) baseline + PnP-ResEB + CC-DuAS (ours) & \textbf{25.92} & \textbf{0.8594} & \textbf{0.0342} \\
\bottomrule
\end{tabularx}
\begin{tablenotes}
\footnotesize
\item $^{*}$ without Residual Pathway.
\end{tablenotes}
% \end{threeparttable}
\label{tab:ablation_tabx}
\end{table*}

\subsection{Ablation Study}
To validate the contribution of each component in the DuSS-PET framework, a comprehensive ablation study was conducted at $\mathrm{DRF}=50$. The results, summarized in Table~\ref{tab:ablation_tabx}, are analyzed as follows:
\begin{itemize}
\item \textbf{A-SwinRB (Baseline)}: Serving as the core sinogram-domain module, the standalone A-SwinRB significantly outperforms the low-dose input, improving PSNR by $+2.5$ dB and SSIM by $0.23$. This underscores the critical importance of direct, self-supervised restoration in the sinogram domain for capturing global structural information.
\item \textbf{+ PnP-ResEB}: Integrating the image-domain prior via the PnP-ResEB module yields the most substantial gains, adding $+3.6$ dB PSNR and $+0.04$ SSIM over the baseline. This demonstrates the indispensable role of a powerful, pre-trained denoiser in recovering fine-grained image details and quantitative accuracy. It is worth noting that the DnCNN prior is pre-trained on \texttt{UDPET-Pretrain} (United Imaging, 2023), which differs in acquisition year from \texttt{UDPET-Main} (United Imaging, 2022), introducing a mild distributional shift. The stable contribution of PnP-ResEB here, together with the consistently strong results in Tables~\ref{tab:qua} and~\ref{tab:gen}, confirms that the multi-DRF Nr2N training regimen yields a generalizable image prior that remains effective under moderate data distribution shifts. Its behavior under substantially larger pretraining-to-reconstruction shifts remains to be further evaluated.
\item \textbf{+ CC-DuAS}: Enforcing consistency through the CC-DuAS strategy provides further, consistent improvements ($+1$ dB PSNR, $+0.01$ SSIM). This indicates that cross-representation consistency serves as an effective self-supervised regularization, improving reconstruction stability and robustness under severe photon sparsity.
\item \textbf{Residual Fusion}: Removing the residual pathway results in a measurable performance drop across all metrics, verifying that this design effectively preserves critical information from the initial reconstruction and mitigates potential over-smoothing.
\item \textbf{Full Model (A-SwinRB + PnP-ResEB + CC-DuAS)}: The complete DuSS-PET framework achieves the best performance, demonstrating that the proposed modules are synergistic and complementary. The full model delivers superior quantitative results and the highest visual fidelity.
\end{itemize}

Overall, the ablation results confirm that each component addresses a distinct yet complementary aspect of the reconstruction problem. A-SwinRB provides a strong initial reconstruction, PnP-ResEB enhances structural details via an image prior, and CC-DuAS stabilizes training by enforcing cross-representation consistency. Together, these elements are essential for achieving high-quality self-supervised PET reconstruction.

\section{Discussion} \label{sec:5}
\subsection{Validation of the Dose-Aware Bernoulli Corruption Strategy}

A core innovation of our work is the Do-ABC strategy, which adapts the Nr2N paradigm for low-dose PET reconstruction. Unlike conventional approaches that use Poisson noise to simulate denoising tasks~\cite{krullNoise2voidlearningDenoisingSingle2019,moranNoisier2noiseLearningDenoise2020, lehtinenNoise2NoiseLearningImage2018a}, our method employs Bernoulli masking to introduce structured information dropout that reflects the severe sparsity commonly observed in low-dose sinogram data. This approach is conceptually aligned with N2V, as the masking operation forces the network to learn robust spatial context for predicting missing information.

To quantitatively validate Do-ABC, we conducted a controlled analysis within the A-SwinRB module, replacing it with a standard Poisson noise model ($\lambda=25$) while keeping all other parameters identical. As summarized in Table~\ref{tab:corruption}, our strategy delivers consistent and significant improvements across all metrics. These results demonstrate that Bernoulli masking provides a more effective self-supervised training signal than Poisson noise under the same experimental setting. By better matching the statistical characteristics of severe signal sparsity, the Do-ABC strategy provides a stronger learning signal for the self-supervised reconstruction task, leading to substantial gains in output quality.

\begin{table}[htbp]
\small
\centering
\caption{Quantitative evaluation of noise corruption strategies in the A-SwinRB module.}
\begin{tabularx}{\linewidth}{lXXX}
\toprule
Method & PSNR & SSIM & NMSE \\
\midrule
LD-PET                       & 19.65 & 0.5700 & 0.1438 \\
(1) Do-ABC    & \textbf{22.02} & \textbf{0.8144} & \textbf{0.0857} \\
(2) Poisson-based      & 21.37 & 0.8028 & 0.0991 \\
\bottomrule
\end{tabularx}
\label{tab:corruption}
\end{table}

Importantly, these results should not be construed as evidence that Bernoulli masking accurately models PET acquisition physics. Instead, they demonstrate that Do-ABC serves as an empirically calibrated corruption strategy, offering a more effective self-supervised learning signal compared to the standard Poisson-based approach in the current retrospective reconstruction framework.

\subsection{Validation and Sensitivity Analysis of the Pre-Selected Retention Probabilities $\alpha$}
The Do-ABC mechanism is parameterized by a dose-specific retention probability $\alpha$, calibrated to approximate the statistical distribution of real low-dose PET images at a given DRF (Sec.~\ref{sec:AA}). This yields pre-selected retention probabilities of $\alpha=0.8,\,0.6,\,0.4$ for $\mathrm{DRF}=20,\,50,\,100$, respectively. 

\begin{figure*}
	\centering
	\includegraphics[width=\textwidth]{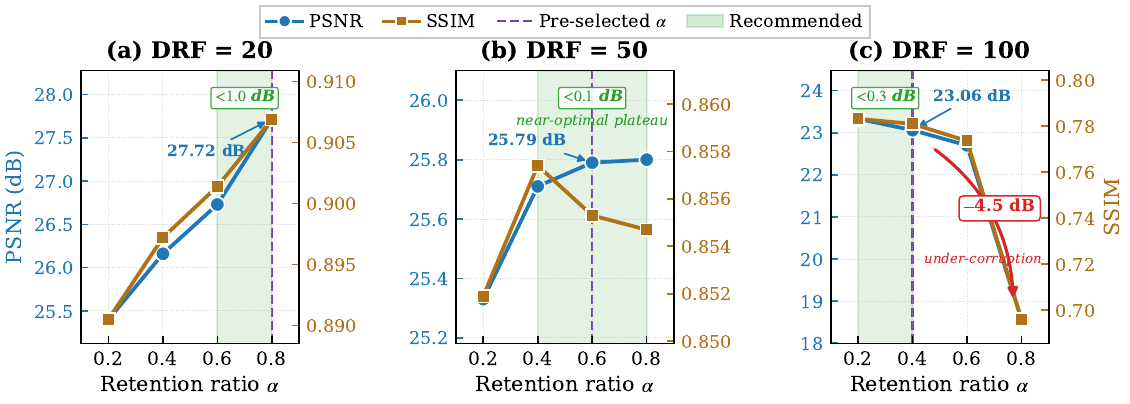}
    \caption{Sensitivity of reconstruction quality to the retention probabilities $\alpha$ at $\mathrm{DRF}=20, 50, 100$. Four retention probabilities ($\alpha=0.2, 0.4, 0.6, 0.8$) are evaluated via PSNR (blue, left axis) and SSIM (brown, right axis). The purple dashed line indicates the pre-selected $\alpha$ for each DRF, as calibrated by the distribution-matching analysis in Fig.~\ref{fig:count}. The green-shaded region denotes the recommended operating range, within which PSNR remains within $1$\,dB of the best observed value. These results validate that the calibrated retention probabilities from Fig.~\ref{fig:count} lie in stable near-optimal regions for $\mathrm{DRF}=20$ and $50$, while sensitivity increases at $\mathrm{DRF}=100$, where under-corruption ($\alpha>0.4$) causes substantial degradation.}

	\label{fig:ab.R}
\end{figure*}

This calibration is grounded in the Nr2N principle~\cite{moranNoisier2noiseLearningDenoise2020}: the secondary corruption intensity should match the intrinsic noise level of the input data. To empirically validate this selection and characterize the sensitivity of reconstruction quality to $\alpha$, we evaluated a range of $\alpha$ values ($0.2,0.4,0.6,0.8$) across all dose levels. As shown in Fig.~\ref{fig:ab.R}, the pre-selected $\alpha$ values consistently lie in near-optimal operating regions across all evaluation metrics for their respective DRFs.

However, the sensitivity profile varies with the dose level. At \mbox{$\mathrm{DRF}=20$}, the method is moderately tolerant: changing the pre-selected $\alpha=0.8$ to $\alpha=0.6$ reduces PSNR by about $1.0$\,dB, while SSIM remains above $0.90$, suggesting stable performance within $\alpha\in[0.6,0.8]$. At \mbox{$\mathrm{DRF}=50$}, the performance surface is relatively flat around the pre-selected $\alpha=0.6$, with PSNR varying by less than $0.1$\,dB across $\alpha\in[0.4,0.8]$, indicating low sensitivity to the retention probabilities at this dose level. In contrast, at \mbox{$\mathrm{DRF}=100$}, the method becomes more sensitive: PSNR decreases by about $4.5$\,dB when $\alpha$ changes from the pre-selected $0.4$ to $0.8$. This trend is consistent with the Nr2N requirement that the secondary corruption should produce a sufficiently noisier counterpart, which becomes harder to satisfy when the input is already severely degraded.

In summary, the pre-selected $\alpha$ values obtained by the distribution-matching calibration in Sec.~\ref{sec:AA} all fall within the stable operating range, while the performance trends under deviations from these values further illustrate the dose-dependent stability of $\alpha$ selection.

\subsection{Limitations and Future Work} \label{sec:limit}

While demonstrating strong performance in 2D low-dose PET reconstruction, the DuSS-PET framework presents three primary limitations that warrant future investigation.

First, the current implementation operates on 2D slices. Recent studies have shown the feasibility of 3D deep learning for PET reconstruction and denoising~\cite{zeng3DCVTGAN3D2022, wang3DMultimodalityTransformerGAN2024, zhaoSelfsupervisedDeepLearning2023}, suggesting that extending DuSS-PET to volumetric modeling is a natural next step. Rather than directly replacing all 2D modules with dense 3D counterparts, a practical implementation would more likely adopt slab- or patch-based designs to balance inter-slice context and computational cost. Under such a setting, the image-domain prior, the A-SwinRB module, and the CC-DuAS mechanism could be reformulated in a lightweight volumetric manner while preserving pseudo-dual-domain consistency. In practice, this extension would likely require memory-efficient training strategies, such as mixed-precision training, gradient checkpointing, and staged optimization. The main challenge is therefore to maintain pseudo-dual-domain consistency under volumetric constraints, rather than merely increasing the network dimensionality.

Second, the Do-ABC mechanism employs a fixed noise injection retention probabilities ($\alpha$) for each dose reduction factor. While effective in controlled settings, this design may limit adaptability to patient-specific or acquisition-dependent noise variations encountered in clinical practice. A promising direction is to integrate an adaptive noise-level estimator to dynamically modulate the corruption process, thereby improving robustness across heterogeneous clinical data.

Third, all sinograms in the current framework are generated by applying a 2D parallel-beam Radon transform to reconstructed PET images, rather than raw measured projections acquired by the scanner. This simplified computational model does not account for several physical factors involved in real PET acquisition, such as three-dimensional geometric effects, photon attenuation, scatter, random coincidences, and detector response. Accordingly, the pseudo-sinogram domain in this study should be regarded as a computational consistency domain rather than a faithful surrogate of the true measurement process. Importantly, the central idea of DuSS-PET, namely self-supervised dual-domain consistency between an image and its projection, is not tied to a specific forward operator. If raw list-mode sinograms or calibrated system-matrix projections become available, the current Radon operator could be replaced with a more realistic system model without changing the overall training paradigm. Evaluating such an extension and its effect on reconstruction fidelity is an important direction for future work.

In summary, while DuSS-PET establishes a robust foundation for unpaired, self-supervised PET reconstruction with pseudo-dual-domain consistency in 2D, future work will focus on bridging the gap to real PET acquisition physics, improving its scalability to 3D, and adaptability to real-world clinical noise profiles.

\section{Conclusion} \label{sec:6}

This paper presented DuSS-PET, a self-supervised PET reconstruction framework that eliminates the need for paired full-dose data by enforcing consistency between reconstructed images and pseudo-sinograms. The proposed method integrates a Bernoulli sampling-based corruption strategy (Do-ABC) with synergistic sinogram restoration (A-SwinRB) and image-domain reconstruction (PnP-ResEB), constrained by a pseudo-dual-domain consistency mechanism (CC-DuAS). Extensive experiments demonstrate that DuSS-PET outperforms existing self-supervised baselines and achieves performance comparable to fully supervised state-of-the-art methods, with strong generalization across scanners. This work establishes a robust and clinically relevant paradigm for unpaired low-dose PET reconstruction. Future work will focus on extending the framework to 3D volumetric reconstruction through module-wise volumetric adaptation and slab-based consistency training.

\appendices
\section{Derivation of Eqs. (3) and (4)}

The derivation follows the multiplicative Bernoulli-noise analysis of Nr2N, adapted to our notation.

Let $I_{\text{FD}}$ denote the latent clean full-dose PET image. Based on the approximation in Sec.~\ref{sec:AA}(2), the first Bernoulli degradation is defined as
\begin{equation}
I_{\text{LD}} \approx I_{\text{De}} = M_{\alpha_0} \odot I_{\text{FD}},
\tag{A1}
\end{equation}
where $M_{\alpha_0} \sim \text{Bernoulli}(\alpha_0)$, and $\alpha_0$ is the retention probability. Therefore, the corresponding masking-to-zero probability is
\[
p = 1 - \alpha_0.
\]

A second Bernoulli corruption is then applied to obtain the noisier image
\begin{equation}
I_{\text{NLD}} = M_{\alpha} \odot I_{\text{LD}},
\tag{A2}
\end{equation}
where $M_{\alpha} \sim \text{Bernoulli}(\alpha)$, and thus the second masking-to-zero probability is
\[
q = 1 - \alpha.
\]

According to the Nr2N analysis for multiplicative Bernoulli noise, if $Z$ denotes the doubly corrupted observation and $Y$ denotes the singly corrupted one, then
\begin{equation}
\mathbb{E}[X \mid Z] = \frac{\mathbb{E}[Y \mid Z] - kZ}{1 - k},
\qquad k = \frac{p}{p + q - pq}.
\tag{A3}
\end{equation}

Using the variable correspondence
\[
X \leftrightarrow I_{\text{FD}}, \quad
Y \leftrightarrow I_{\text{LD}}, \quad
Z \leftrightarrow I_{\text{NLD}},
\]
and substituting $p = 1 - \alpha_0$, $q = 1 - \alpha$, we obtain
\begin{equation}
k = \frac{1 - \alpha_0}
         {(1\!-\!\alpha_0) + (1\!-\!\alpha) - (1\!-\!\alpha_0)(1\!-\!\alpha)}.
\tag{A4}
\end{equation}
Expanding the denominator gives
\begin{align*}
&(1\!-\!\alpha_0) + (1\!-\!\alpha)
  - (1\!-\!\alpha_0\!-\!\alpha\!+\!\alpha_0 \alpha)\notag\\
&\quad= 1 - \alpha_0 \alpha.
\end{align*}
Hence,
\begin{equation}
k = \frac{1 - \alpha_0}{1 - \alpha_0 \alpha}.
\tag{A5}
\end{equation}
Therefore,
\begin{equation}
\mathbb{E}[I_{\text{FD}} \mid I_{\text{NLD}}] = \frac{\mathbb{E}[I_{\text{LD}} \mid I_{\text{NLD}}] - k I_{\text{NLD}}}{1 - k},
\tag{A6}
\end{equation}
which yields Eq.~(3).

For Eq.~(4), let $\mathcal{FP}(\cdot)$ denote the 2D forward-projection operator. Since $\mathcal{FP}(\cdot)$ is linear, conditional expectation commutes with the linear transform in this setting, giving
\begin{multline*}
\mathbb{E}[\mathcal{FP}(I_{\text{FD}}) \mid \mathcal{FP}(I_{\text{NLD}})]\\
  = \frac{\mathbb{E}[\mathcal{FP}(I_{\text{LD}}) \mid \mathcal{FP}(I_{\text{NLD}})]
         - k\,\mathcal{FP}(I_{\text{NLD}})}{1 - k}.
\end{multline*}
Defining $S_{\text{FD}}\!=\!\mathcal{FP}(I_{\text{FD}})$,
$S_{\text{LD}}\!=\!\mathcal{FP}(I_{\text{LD}})$, and
$S_{\text{NLD}}\!=\!\mathcal{FP}(I_{\text{NLD}})$,
we obtain
\begin{equation}
\mathbb{E}[S_{\text{FD}} \mid S_{\text{NLD}}] = \frac{\mathbb{E}[S_{\text{LD}} \mid S_{\text{NLD}}] - k S_{\text{NLD}}}{1 - k},
\tag{A7}
\end{equation}
which is Eq.~(4).

\section*{Acknowledgment}
All authors declare that they have no known conflicts of interest in terms of competing financial interests or personal relationships that could have an influence or are relevant to the work reported in this paper.

\bibliographystyle{IEEEtran}
\bibliography{reference} % references.bib 为你的 zotero 导出文件

% Generated by IEEEtran.bst, version: 1.14 (2015/08/26)
\begin{thebibliography}{10}
\providecommand{\url}[1]{#1}
\csname url@samestyle\endcsname
\providecommand{\newblock}{\relax}
\providecommand{\bibinfo}[2]{#2}
\providecommand{\BIBentrySTDinterwordspacing}{\spaceskip=0pt\relax}
\providecommand{\BIBentryALTinterwordstretchfactor}{4}
\providecommand{\BIBentryALTinterwordspacing}{\spaceskip=\fontdimen2\font plus
\BIBentryALTinterwordstretchfactor\fontdimen3\font minus \fontdimen4\font\relax}
\providecommand{\BIBforeignlanguage}[2]{{%
\expandafter\ifx\csname l@#1\endcsname\relax
\typeout{** WARNING: IEEEtran.bst: No hyphenation pattern has been}%
\typeout{** loaded for the language `#1'. Using the pattern for}%
\typeout{** the default language instead.}%
\else
\language=\csname l@#1\endcsname
\fi
#2}}
\providecommand{\BIBdecl}{\relax}
\BIBdecl

\bibitem{apostolovaImagingTumorMetabolism2016}
I.~Apostolova, F.~Wedel, and W.~Brenner, ``Imaging of tumor metabolism using positron emission tomography ({{PET}}),'' \emph{Metab. Cancer}, pp. 177--205, 2016.

\bibitem{zhangComparisonWholebody18FFDG2023}
C.~Zhang, Z.~Liang, W.~Liu, X.~Zeng, and Y.~Mo, ``Comparison of whole-body {{18F-FDG PET}}/{{CT}} and {{PET}}/{{MRI}} for distant metastases in patients with malignant tumors: a meta-analysis,'' \emph{BMC cancer}, vol.~23, no.~1, p.~37, 2023.

\bibitem{spick18FFDGPETCT2016}
C.~Spick, K.~Herrmann, and J.~Czernin, ``{{18F-FDG PET}}/{{CT}} and {{PET}}/{{MRI}} perform equally well in cancer: evidence from studies on more than 2,300 patients,'' \emph{J. Nucl. Med.}, vol.~57, no.~3, pp. 420--430, 2016.

\bibitem{gonzalez-montoroAdvancesDetectorInstrumentation2022}
A.~{Gonzalez-Montoro}, M.~N. Ullah, and C.~S. Levin, ``Advances in detector instrumentation for {{PET}},'' \emph{J. Nucl. Med.}, vol.~63, no.~8, pp. 1138--1144, 2022.

\bibitem{yanMethodAssessImage2016}
J.~Yan, J.~Schaefferkoetter, M.~Conti, and D.~Townsend, ``A method to assess image quality for low-dose {{PET}}: analysis of {{SNR}}, {{CNR}}, bias and image noise,'' \emph{Cancer Imaging}, vol.~16, pp. 1--12, 2016.

\bibitem{watsonOptimizingInjectedDose2005}
C.~C. Watson, M.~E. Casey, B.~Bendriem, J.~P. Carney, D.~W. Townsend, S.~Eberl, S.~Meikle, and F.~P. DiFilippo, ``Optimizing injected dose in clinical {{PET}} by accurately modeling the counting-rate response functions specific to individual patient scans,'' \emph{J. Nucl. Med.}, vol.~46, no.~11, pp. 1825--1834, 2005.

\bibitem{zhaoImageQualityLesion2021}
Y.-M. Zhao, Y.-H. Li, T.~Chen, W.-G. Zhang, L.-H. Wang, J.~Feng, C.~Li, X.~Zhang, W.~Fan, and Y.-Y. Hu, ``Image quality and lesion detectability in low-dose pediatric 18 f-{{FDG}} scans using total-body {{PET}}/{{CT}},'' \emph{Eur. J. Nucl. Med. Mol. Imaging}, vol.~48, pp. 3378--3385, 2021.

\bibitem{ramachandran1971three}
{\relax GN}.~Ramachandran and {\relax AV}.~Lakshminarayanan, ``Three-dimensional reconstruction from radiographs and electron micrographs: application of convolutions instead of {{Fourier}} transforms,'' \emph{Proc. Natl. Acad. Sci.}, vol.~68, no.~9, pp. 2236--2240, 1971.

\bibitem{hudsonAcceleratedImageReconstruction1994}
H.~M. Hudson and R.~S. Larkin, ``Accelerated image reconstruction using ordered subsets of projection data,'' \emph{IEEE Trans. Med. Imaging}, vol.~13, no.~4, pp. 601--609, 1994.

\bibitem{readerAIPETImage2023}
A.~J. Reader and B.~Pan, ``{AI} for {PET} image reconstruction,'' \emph{Br. J. Radiol.}, vol.~96, no. 1150, p. 20230292, 2023.

\bibitem{xiangDeepAutocontextConvolutional2017}
L.~Xiang, Y.~Qiao, D.~Nie, L.~An, W.~Lin, Q.~Wang, and D.~Shen, ``Deep auto-context convolutional neural networks for standard-dose {{PET}} image estimation from low-dose {{PET}}/{{MRI}},'' \emph{Neurocomputing}, vol. 267, pp. 406--416, 2017.

\bibitem{wangLocalityAdaptiveMultimodality2018}
Y.~Wang, L.~Zhou, L.~Wang, B.~Yu, C.~Zu, D.~S. Lalush, W.~Lin, X.~Wu, J.~Zhou, and D.~Shen, ``Locality adaptive multi-modality gans for high-quality {{PET}} image synthesis,'' in \emph{Med. {{Image Comput}}. {{Comput}}. {{Assist}}. {{Interv}}. 2018 21st {{Int}}. {{Conf}}. {{Granada Spain Sept}}. 16-20 2018 {{Proc}}. {{Part I}}}.\hskip 1em plus 0.5em minus 0.4em\relax Springer, 2018, pp. 329--337.

\bibitem{zhouSupervisedLearningCyclegan2020}
L.~Zhou, J.~D. Schaefferkoetter, I.~W. Tham, G.~Huang, and J.~Yan, ``Supervised learning with cyclegan for low-dose {{FDG PET}} image denoising,'' \emph{Med. Image Anal.}, vol.~65, p. 101770, 2020.

\bibitem{zeng3DCVTGAN3D2022}
P.~Zeng, L.~Zhou, C.~Zu, X.~Zeng, Z.~Jiao, X.~Wu, J.~Zhou, D.~Shen, and Y.~Wang, ``{{3D CVT-GAN}}: a {{3D}} convolutional vision transformer-{{GAN}} for {{PET}} reconstruction,'' in \emph{Medical Image Computing and Computer Assisted Intervention -- {MICCAI} 2022}.\hskip 1em plus 0.5em minus 0.4em\relax Springer, 2022, pp. 516--526.

\bibitem{shenPETImageDenoising2023}
C.~Shen, Z.~Yang, and Y.~Zhang, ``{PET} image denoising with score-based diffusion probabilistic models,'' in \emph{Medical Image Computing and Computer Assisted Intervention -- {MICCAI} 2023}.\hskip 1em plus 0.5em minus 0.4em\relax Springer, 2023, pp. 270--278.

\bibitem{sanaatProjectionSpaceImplementation2020}
A.~Sanaat, H.~Arabi, I.~Mainta, V.~Garibotto, and H.~Zaidi, ``Projection space implementation of deep learning-guided low-dose brain {PET} imaging improves performance over implementation in image space,'' \emph{J. Nucl. Med.}, vol.~61, no.~9, pp. 1388--1396, 2020.

\bibitem{whiteleyDirectPETFullsizeNeural2020}
W.~Whiteley, W.~K. Luk, and J.~Gregor, ``Directpet: full-size neural network {PET} reconstruction from sinogram data,'' \emph{J. Med. Imaging}, vol.~7, no.~3, p. 032503, 2020.

\bibitem{haggstromDeepPET2019}
I.~H{\"a}ggstr{\"o}m, C.~R. Schmidtlein, G.~Campanella, and T.~J. Fuchs, ``Deeppet: a deep encoder-decoder network for directly solving the {PET} image reconstruction inverse problem,'' \emph{Med. Image Anal.}, vol.~54, pp. 253--262, 2019.

\bibitem{aiREDResidualEstimation2025}
X.~Ai, B.~Huang, F.~Chen, L.~Shi, B.~Li, S.~Wang, and Q.~Liu, ``{RED}: Residual estimation diffusion for low-dose {PET} sinogram reconstruction,'' \emph{Med. Image Anal.}, vol. 102, p. 103558, 2025.

\bibitem{liuDeeplearningbasedFrameworkPET2022}
Z.~Liu, H.~Ye, and H.~Liu, ``Deep-learning-based framework for {{PET}} image reconstruction from sinogram domain,'' \emph{Appl. Sci.}, vol.~12, no.~16, p. 8118, 2022.

\bibitem{dossUtilizingDeepLearning2023}
K.~K. Manoj~Doss and J.-C. Chen, ``Utilizing deep learning techniques to improve image quality and noise reduction in preclinical low-dose {PET} images in the sinogram domain,'' \emph{Med. Phys.}, vol.~51, no.~1, pp. 209--223, 2024.

\bibitem{huTransEMResidualSwintransformer2022}
R.~Hu and H.~Liu, ``Transem: Residual swin-transformer based regularized {PET} image reconstruction,'' in \emph{Medical Image Computing and Computer Assisted Intervention -- {MICCAI} 2022}.\hskip 1em plus 0.5em minus 0.4em\relax Springer, 2022, pp. 184--193.

\bibitem{mehranianModelbasedDeepLearning2020}
A.~Mehranian and A.~J. Reader, ``Model-based deep learning {{PET}} image reconstruction using forward-backward splitting expectation-maximization,'' \emph{IEEE Trans. Radiat. Plasma Med. Sci.}, vol.~5, no.~1, pp. 54--64, 2020.

\bibitem{cuiTriDoFormerTripleDomainTransformer2023}
J.~Cui, P.~Zeng, X.~Zeng, P.~Wang, X.~Wu, J.~Zhou, Y.~Wang, and D.~Shen, ``Trido-former: A triple-domain transformer for direct {PET} reconstruction from low-dose sinograms,'' in \emph{Medical Image Computing and Computer Assisted Intervention -- {MICCAI} 2023}.\hskip 1em plus 0.5em minus 0.4em\relax Springer, 2023, pp. 184--194.

\bibitem{cuiPriorKnowledgeguidedTripleDomain2024}
J.~Cui, P.~Zeng, X.~Zeng, Y.~Xu, P.~Wang, J.~Zhou, Y.~Wang, and D.~Shen, ``Prior knowledge-guided triple-domain transformer-gan for direct {PET} reconstruction from low-count sinograms,'' \emph{IEEE Trans. Med. Imaging}, vol.~43, no.~12, pp. 4174--4189, 2024.

\bibitem{zhaoSelfsupervisedDeepLearning2023}
F.~Zhao, D.~Li, R.~Luo, M.~Liu, X.~Jiang, and J.~Hu, ``Self-supervised deep learning for joint {{3D}} low-dose {{PET}}/{{CT}} image denoising,'' \emph{Computers in Biology and Medicine}, vol. 165, p. 107391, 2023.

\bibitem{arabiNoisetonoiseModelTraining2022}
H.~Arabi and H.~Zaidi, ``Noise-to-noise model training for noise reduction in {{PET}} imaging using multiple image reconstructions,'' in \emph{2022 {{IEEE Nucl}}. {{Sci}}. {{Symp}}. {{Med}}. {{Imaging Conf}}. {{NSSMIC}}}.\hskip 1em plus 0.5em minus 0.4em\relax IEEE, 2022, pp. 1--3.

\bibitem{lehtinenNoise2NoiseLearningImage2018a}
J.~Lehtinen, J.~Munkberg, J.~Hasselgren, S.~Laine, T.~Karras, M.~Aittala, and T.~Aila, ``Noise2noise: Learning image restoration without clean data,'' in \emph{Proceedings of the 35th International Conference on Machine Learning}.\hskip 1em plus 0.5em minus 0.4em\relax PMLR, 2018, pp. 2965--2974.

\bibitem{moranNoisier2noiseLearningDenoise2020}
N.~Moran, D.~Schmidt, Y.~Zhong, and P.~Coady, ``Noisier2noise: Learning to denoise from unpaired noisy data,'' in \emph{Proceedings of the IEEE/CVF Conference on Computer Vision and Pattern Recognition ({CVPR})}, 2020, pp. 12\,064--12\,072.

\bibitem{xueCrossscannerCrosstracerDeep2022}
S.~Xue, R.~Guo, K.~P. Bohn, J.~Matzke, M.~Viscione, I.~Alberts, H.~Meng, C.~Sun, M.~Zhang, M.~Zhang, R.~Sznitman, G.~El~Fakhri, A.~Rominger, B.~Li, and K.~Shi, ``A cross-scanner and cross-tracer deep learning method for the recovery of standard-dose imaging quality from low-dose {PET},'' \emph{Eur. J. Nucl. Med. Mol. Imaging}, vol.~49, pp. 1843--1856, 2022.

\bibitem{fallahpoorDeepLearningTechniques2024}
M.~Fallahpoor, S.~Chakraborty, B.~Pradhan, O.~Faust, P.~D. Barua, H.~Chegeni, and R.~Acharya, ``Deep learning techniques in {{PET}}/{{CT}} imaging: a comprehensive review from sinogram to image space,'' \emph{Comput. Methods Programs Biomed.}, vol. 243, p. 107880, 2024.

\bibitem{hashimotoDeepLearningBasedPET2024}
F.~Hashimoto, Y.~Onishi, K.~Ote, H.~Tashima, A.~J. Reader, and T.~Yamaya, ``Deep learning-based {PET} image denoising and reconstruction: a review,'' \emph{Radiol. Phys. Technol.}, vol.~17, no.~1, pp. 24--46, 2024.

\bibitem{zhangGeneralizableUnpairedPET2026}
Y.~Luo, J.~Cui, H.~Zhang, Y.~Yang, H.~Shalihaer, Y.~Feng, X.~Su, H.~Liu, and X.~Li, ``Generalizable unpaired {PET} image enhancement via two-stage diffusion framework,'' \emph{IEEE Trans. Radiat. Plasma Med. Sci.}, pp. 1--1, 2026.

\bibitem{gongPETImageDenoisingDDPM2024}
K.~Gong, K.~Johnson, G.~El~Fakhri, Q.~Li, and T.~Pan, ``{PET} image denoising based on denoising diffusion probabilistic model,'' \emph{Eur. J. Nucl. Med. Mol. Imaging}, vol.~51, no.~2, pp. 358--368, 2024.

\bibitem{krullNoise2voidlearningDenoisingSingle2019}
A.~Krull, T.-O. Buchholz, and F.~Jug, ``Noise2void -- learning denoising from single noisy images,'' in \emph{Proceedings of the IEEE/CVF Conference on Computer Vision and Pattern Recognition ({CVPR})}, 2019, pp. 2129--2137.

\bibitem{batsonNoise2selfBlindDenoising2019}
J.~Batson and L.~Royer, ``Noise2self: {{Blind}} denoising by self-supervision,'' in \emph{Int. {{Conf}}. {{Mach}}. {{Learn}}.}\hskip 1em plus 0.5em minus 0.4em\relax PMLR, 2019, pp. 524--533.

\bibitem{bousseReviewLowDoseEmission2024}
A.~Bousse, V.~S.~S. Kandarpa, K.~Shi, K.~Gong, J.~S. Lee, C.~Liu, and D.~Visvikis, ``A review on low-dose emission tomography post-reconstruction denoising with neural network approaches,'' \emph{IEEE Trans. Radiat. Plasma Med. Sci.}, vol.~8, no.~4, pp. 333--347, 2024.

\bibitem{yamanSelfsupervisedLearningPhysicsguided2020}
B.~Yaman, S.~A.~H. Hosseini, S.~Moeller, J.~Ellermann, K.~U{\u g}urbil, and M.~Ak{\c c}akaya, ``Self-supervised learning of physics-guided reconstruction neural networks without fully sampled reference data,'' \emph{Magn. Reson. Med.}, vol.~84, no.~6, pp. 3172--3191, 2020.

\bibitem{millardTheoreticalFrameworkSelfsupervised2023}
C.~Millard and M.~Chiew, ``A theoretical framework for self-supervised {{MR}} image reconstruction using sub-sampling via variable density {{Noisier2Noise}},'' \emph{IEEE Trans. Comput. Imaging}, vol.~9, pp. 707--720, 2023.

\bibitem{yieSelfsupervisedPETDenoising2020}
S.~Y. Yie, S.~K. Kang, D.~Hwang, and J.~S. Lee, ``Self-supervised {PET} denoising,'' \emph{Nucl. Med. Mol. Imaging}, vol.~54, no.~6, pp. 299--304, 2020.

\bibitem{yieEffectsLossFunctions2024}
S.~Y. Yie, K.~M. Kim, S.~Bae, and J.~S. Lee, ``Effects of loss functions and supervision methods on total-body {PET} denoising,'' \emph{IEEE Trans. Radiat. Plasma Med. Sci.}, vol.~8, no.~4, pp. 379--390, 2024.

\bibitem{songSelfsupervisedPETImage2023}
T.-A. Song, F.~Yang, and J.~Dutta, ``Self-supervised {PET} image denoising using a neighbor-to-neighbor network,'' \emph{J. Nucl. Med.}, vol.~64, no. supplement 1, p. P1380, 2023.

\bibitem{liNeuralArchitectureSearchUnsupervised2026}
J.~Li, J.~Wang, Y.~Lv, P.~Zhang, and J.~Zhao, ``Neural architecture search for unsupervised {PET} image denoising,'' \emph{IEEE Trans. Radiat. Plasma Med. Sci.}, vol.~10, no.~1, pp. 51--62, 2026.

\bibitem{xieUnifiedNoiseAwareNetwork2024}
H.~Xie, Q.~Liu, B.~Zhou, X.~Chen, X.~Guo, H.~Wang, B.~Li, A.~Rominger, K.~Shi, and C.~Liu, ``Unified noise-aware network for low-count {PET} denoising with varying count levels,'' \emph{IEEE Trans. Radiat. Plasma Med. Sci.}, vol.~8, no.~4, pp. 366--378, 2024.

\bibitem{songNoise2VoidUnsupervisedDenoising2021}
T.-A. Song, F.~Yang, and J.~Dutta, ``Noise2void: unsupervised denoising of {PET} images,'' \emph{Phys. Med. Biol.}, vol.~66, no.~21, p. 214002, 2021.

\bibitem{liuPersonalizedDeepLearning2022}
Q.~Liu, H.~Liu, N.~Mirian, S.~Ren, V.~Viswanath, J.~Karp, S.~Surti, and C.~Liu, ``A personalized deep learning denoising strategy for low-count {PET} images,'' \emph{Phys. Med. Biol.}, vol.~67, no.~14, p. 145014, 2022.

\bibitem{seyyediDeepLearningbasedTechniques2024a}
N.~Seyyedi, A.~Ghafari, N.~Seyyedi, and P.~Sheikhzadeh, ``Deep learning-based techniques for estimating high-quality full-dose positron emission tomography images from low-dose scans: a systematic review,'' \emph{BMC Med. Imaging}, vol.~24, no.~1, p. 238, 2024.

\bibitem{hansenDiscreteInverseProblems2010}
P.~C. Hansen, \emph{Discrete inverse problems: insight and algorithms}.\hskip 1em plus 0.5em minus 0.4em\relax SIAM, 2010.

\bibitem{liangSwinirImageRestoration2021}
J.~Liang, J.~Cao, G.~Sun, K.~Zhang, L.~Van~Gool, and R.~Timofte, ``Swinir: Image restoration using swin transformer,'' in \emph{Proceedings of the IEEE/CVF International Conference on Computer Vision Workshops ({ICCVW})}, 2021, pp. 1833--1844.

\bibitem{zhangGaussianDenoiserResidual2017}
K.~Zhang, W.~Zuo, Y.~Chen, D.~Meng, and L.~Zhang, ``Beyond a gaussian denoiser: {{Residual}} learning of deep {{CNN}} for image denoising,'' \emph{IEEE Trans. Image Process.}, vol.~26, no.~7, pp. 3142--3155, 2017.

\bibitem{zhouDSFormerDualdomainSelfsupervised2023}
B.~Zhou, N.~Dey, J.~Schlemper, S.~S.~M. Salehi, C.~Liu, J.~S. Duncan, and M.~Sofka, ``{{DSFormer}}: a dual-domain self-supervised transformer for accelerated multi-contrast {{MRI}} reconstruction,'' in \emph{Proc. {{IEEECVF Winter Conf}}. {{Appl}}. {{Comput}}. {{Vis}}.}, 2023, pp. 4966--4975.

\bibitem{jangSpachTransformerSpatial2024}
S.-I. Jang, T.~Pan, Y.~Li, P.~Heidari, J.~Chen, Q.~Li, and K.~Gong, ``Spach transformer: {{Spatial}} and channel-wise transformer based on local and global self-attentions for {{PET}} image denoising,'' \emph{IEEE Trans. Med. Imaging}, vol.~43, pp. 2036--2049, 2024.

\bibitem{ronchettiTorchRadonFastDifferentiable2020}
M.~Ronchetti, ``{{TorchRadon}}: {{Fast}} differentiable routines for computed tomography,'' \emph{ArXiv Prepr. ArXiv200914788}, 2020.

\bibitem{wang3DMultimodalityTransformerGAN2024}
Y.~Wang, Y.~Luo, C.~Zu, B.~Zhan, Z.~Jiao, X.~Wu, J.~Zhou, D.~Shen, and L.~Zhou, ``{{3D}} multi-modality {{Transformer-GAN}} for high-quality {{PET}} reconstruction,'' \emph{Medical Image Analysis}, vol.~91, p. 102983, 2024.

\end{thebibliography}

\end{document}